\documentclass[bsc,twoside,parskip,deptreport]{infthesis}
\course{Computer Science}
\project{Fourth Year Project Report}
\usepackage[utf8]{inputenc}
\usepackage{graphicx}
\usepackage{subcaption}
\usepackage{adjustbox}
\usepackage{url}
\usepackage{siunitx}
\usepackage{amsmath}
\usepackage{geometry}
\usepackage{eushield}
\usepackage{ifthen}
\DeclareMathOperator{\csch}{csch}
\usepackage{changepage}
\usepackage{hyperref}
\hypersetup{
    colorlinks=true,
    linkcolor=black,
    filecolor=black,      
    urlcolor=black,
    citecolor=black,
}
\date{January 2018}

\title{Dynamic Walking of Legged Machines}
\author{Kendeas Theofanous}
\submityear{2018}

\abstract{%
Locomotion of legged machines faces the problems of model complexity and computational costs. Algorithms based on complex models and/or reinforcement learning exist to solve the walking control task. In this project, we aim to develop a bipedal walking control system based on a simple model the Linear Inverted Pendulum model. In order to simplify the complex process of controlling legged locomotion, we make use of the technique of splitting the control into three parts as height control, forward velocity control and balance control. The forward velocity of the body has a linear relationship with the foot placement, therefore we use a linear function to realise foot placement. Our control system achieves stable walking gait in a simulated environment, where our bipedal robot walks more than 200 steps with a cyclic pattern in a stable, dynamic and almost natural manner. The experimental data are presented and analysed.
}

\begin{document}

\begin{preliminary}

\maketitle

\begin{acknowledgements}
I would like to thank my supervisor, Dr. Zhibin Li, for his patience, guidance and help. I would also like to thank Chuanyu Yang, member of the Edinburgh Centre of Robotics, for his suggestions and help in proper utilisation of the simulated environment used in the project.
\end{acknowledgements}

\standarddeclaration

\tableofcontents
\listoffigures
\listoftables

\end{preliminary}
\chapter{Introduction}
Robots have been part of society, mainly as industrial machines in production lines, for many years now. Legged machines and natural locomotion is a more interesting research area though. Choosing legs for robot locomotion has various advantages compared to wheels, even though wheels usually provide lower energy efficiency and simpler control. Wheeled robots are the perfect solution for tasks on road and even terrain, but do not perform well on rough terrain. Legged robots can traverse through soft or uneven terrain, which are places inaccessible to wheeled machines and do tasks that are too dangerous for humans. The ideal scenario for the future, that many researchers share, is to have some sort of hybrid machines being able to traverse through terrain with wheels as much as possible and when the terrain does not allow it to be able to use legs to do tasks, thus exploiting the benefits of both methods of transportation.

Legged machines researchers face the problems of model complexity and computational cost. Control of such machines requires coordination of many components when traversing through terrain. Joints control, balance, foot placement as well as understanding of the dynamics of the robot are essential for any legged robot action, whether that be standing, walking, running or jumping.

Bipedal machines are the subject of humanoid robotics. For humanoid robots to be integrated successfully in modern society, they need to be able to do tasks humans do not want to do themselves. In order to achieve this, humanoid robots need to be able to operate in human environment, use tools like humans and have similar shape as humans. Assuming we build robots in human-like shape, dynamic bipedal walking approaches must achieve reliable and natural locomotion of the robot in human environment and taking it a step further, handling tools. 

In this project, we are focusing on bipedal walking. Controlling a bipedal robot and achieving stable locomotion is a challenging task, considering the large number of degrees of freedom (DOF) required for a human-like robot as well as all the mechanical components and joints necessary to build it. This complicated task is simplified, using simple models to explain and generate walking.

In order to simplify the walking task, we must first realise what is walking. The concise definition of the word "walk" is given by the Oxford Advanced Learner’s Dictionary as

\begin{adjustwidth}{1cm}{1cm}
  \textbf{Walk}:\\
    move along at a moderate pace by lifting up and putting down each foot in
    turn, so that one foot is on the ground while the other is being lifted
    (Oxford Advanced Learner’s Dictionary, Oxford University Press)
\end{adjustwidth}

There are two types of walking, static and dynamic. Static walking robots are always balanced, meaning during walking, their CoM is always inside the support polygon. Dynamic walking robots are not always balanced and their CoM leaves the support polygon during some phases of walking.

Now that we have defined walking and its two types, we need to decide how to approach the walking task. We are interested in natural and dynamic walking. In recent years, reinforcement learning (RL) algorithms are getting the spotlight of attention as they study how an agent can learn how to achieve goals in a complex, uncertain environment by interacting with it and observing the results of those interactions. The robot learns how to make decisions on its own for controlling its motors. RL though faces a couple of challenges, like the curse of dimensionality, the curse of real-world samples and the curse of under-modelling and model uncertainty \cite{DBLP:journals/ijrr/KoberBP13}.

In order to start tackling the walking task in a simple manner, without considering complex and computationally expensive RL algorithms, we need to simplify the structure of the robot. Inverted pendulum dynamics demonstrates similarities to walking dynamics. There exist a couple of inverted pendulum models, but we will focus on the Linear Inverted Pendulum model \cite{DBLP:journals/jrm/KajitaT93}. This model offers the advantage of a simple robot structure and a linear function to model the robot's dynamics. 

The next move is to decide on the control of the robot to carry the model's dynamics. Raibert introduced a simple three-part intuitive controller for a one-legged hopper \cite{Raibert_book}. His technique of decomposing the complicated task of walking into separate distinct controllers was heavily used since then. Many used this method for bipedal walking tasks as well \cite{DBLP:phd/ndltd/Pratt00}, \cite{DBLP:conf/iros/YouLCT15}, \cite{DBLP:conf/icra/PrattP98}. We will use the same method as well and decompose the control into height, forward speed and balance control. In this project, we will implement our controllers and run experiments in a simulated environment only.

The remaining of this project is organised as follows: The necessary background knowledge, involving LIP models and control approaches, is discussed in Chapter 2. The proposed methodology is presented in Chapter 3, along with the ideal system design and the walking approach used. Chapter 4 shows our experimentation in the simulated environment and our results. The project ends with a conclusion and an outlook to future research.

Contributions:
\begin{itemize}
    \item Discussion of Linear Inverted Pendulum Model (LIPM), both in 2D and 3D, showing equations for displacement and velocity of CoM.
    \item Discussion of some natural dynamic mechanisms, that could be used to simplify the walking task. These mechanisms were not ideal for our simulated environment, but the study was an inspiration for future work.
    \item Utilisation of existing code in simulation for a four-joint robot, which made use of LIPM in 2-D. 
    \item Our own implementation for robot control for a four-joint robot and extended to a six-joint robot, also in 2-D.  
\end{itemize}

\newpage
\chapter{Background}

\section{Definitions}
In order to go deeper into the models, some fundamental definitions must be understood.

\textbf{Centre of Mass (CoM):}
It is the point which represents the average position of an object, weighted according to their masses. Using the centre of mass we can solve complex problems, by treating objects as if their whole mass is concentrated at that point. 

\textbf{Zero Moment Point (ZMP):}
It is a concept introduced by Vukobratović et al. and specifies the point where no moment in the horizontal direction is produced with respect to (w.r.t) reaction forces when the foot is in contact with the ground. 

\section{Inverted Pendulum Models}
Simple planar models of bipedal walking reduce the complexity of walking algorithms and help us draw intuition from them. The models described in this section are based on Inverted Pendulum whose dynamics are similar to bipedal walking. Using these models, we can treat a robot as a single point mass having mass-less legs.

\subsection{Linear Inverted Pendulum}
The Linear Inverted Pendulum Model (LIPM) was introduced by Kajita and Tani in 1991 \cite{DBLP:journals/jrm/KajitaT93}. In some walking algorithms, LIPM is used to decouple the sagittal and the lateral walking direction. In those walking algorithms, the basic assumption is that the two walking directions do not interfere, so that decoupling does not influence the model accuracy. Then, the two axis of movement are considered separately with their own pendulum as reference.

\subsubsection{2D Linear Inverted Pendulum}\label{2dLIPMsection}
\begin{figure}[ht]
    \includegraphics[width=8cm,height=4cm]{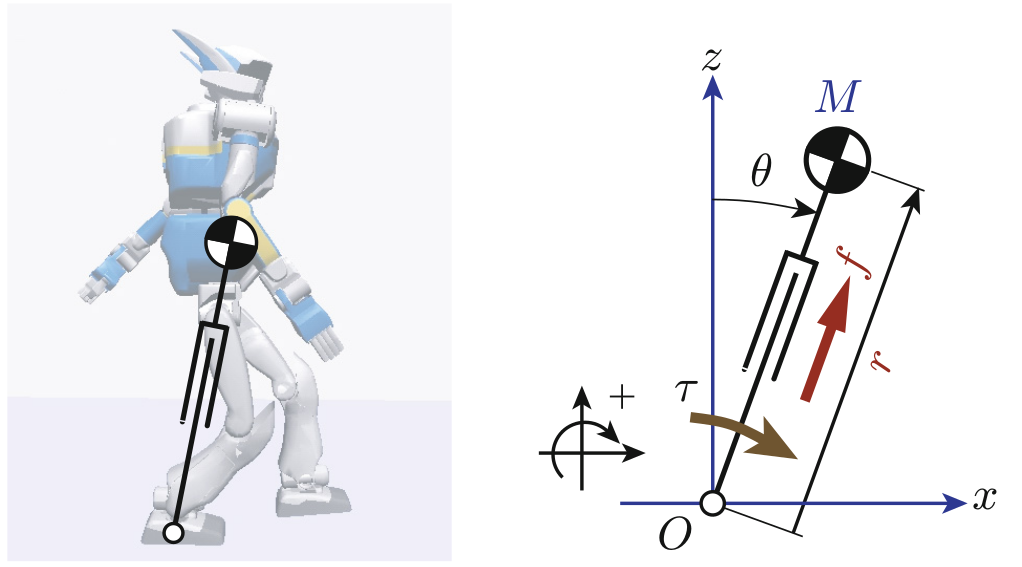}
    \centering
    \caption{2D Inverted pendulum view on a robot and on xz space \cite{intro_book}.}
    \label{fig:2d}
\end{figure}

In this project, we will exploit this assumption and model a biped robot as a 2D inverted pendulum, focusing on the sagittal plane as shown in Figure \ref{fig:2d}. Another assumption necessary is that the robot is only supported by one leg at all times, with the other being in motion. As the whole mass of the robot is at the body, the CoM of the whole robot is the CoM of the body. 

To determine the motion of the CoM, a kick force \textit{f} is required, which when varied gives different falling patterns. We are interested in the kick force
\begin{equation}
    \label{importantf}
    f = \frac{Mg}{\cos \theta}.
\end{equation}
In this case, the CoM moves horizontally, keeping a constant height. This is due to the cancellation of the vertical component by gravity. The horizontal component remains, accelerating the CoM in the horizontal axis. Thus, the motion can be described as:
\begin{equation}
    \label{CoMmotion}
    {M\ddot{x}} = f{\sin \theta}.
\end{equation}
By substituting the kick force \textit{f} from Equation \ref{importantf} into Equation \ref{CoMmotion} we end up with a differential equation describing the dynamics of the CoM in the horizontal axis as:
\begin{equation}
    \ddot{x} = \frac{g}{z}x
\end{equation}
where x, z give the CoM of the inverted pendulum.

As we are interested in keeping a constant height of the CoM, \textit{z} is a constant. Therefore, by solving the differential equation above, equations for the displacement and velocity of the CoM can be expressed as:
\begin{equation}
    x(t) = x(0) \cosh(\frac{t}{T_c}) + T_c\dot{x}(0) \sinh(\frac{t}{T_c})
\end{equation}
\begin{equation}
    \dot{x}(t) = \frac{x(0)}{T_c} \sinh(\frac{t}{T_c}) + \dot{x}(0) \cosh(t/T_c)
\end{equation}
\begin{equation}
    T_c = \sqrt{\frac{z}{g}}
\end{equation}
where \textit{T$_c$} is a time constant given by the constant height \textit{z} and gravity.

\subsubsection{Extension of LIPM for uneven terrain}
So far, the equations explained make walking possible on flat terrain. A similar approach can be used to achieve walking on uneven terrain, without escaping the simplicity of the LIPM. If we constrain the CoM to move on a sloped straight line, the motion is described by:
\begin{equation}
    y = kx + y_c
\end{equation}
where \textit{k} is the slope of the \textit{constraint line} and \textit{y$_c$} is the intersection with the \textit{y}-axis, which must be greater than zero to serve our purposes.

Calculating the kick force \textit{f} horizontal and vertical components and forcing the sum of \textit{f} and gravity to be parallel to the constraint line, the following equation is obtained:
\begin{equation}
    \label{fmgr}
    f = \frac{Mgr}{y_c}
\end{equation}

Similar to the constant height LIPM, gravity is cancelled by the kick force and the horizontal component of \textit{f} accelerates the CoM. In order to obtain the horizontal dynamics of the CoM, we use Equation \ref{fmgr} and the horizontal component of \textit{f} obtaining an identical equation as the constant height LIPM:
\begin{equation}
    \ddot{x} = \frac{g}{z_c}x
\end{equation}

Therefore, walking patterns can be designed for traversal on uneven terrain. 
\begin{figure}[ht]
    \includegraphics[width=5cm,height=4cm]{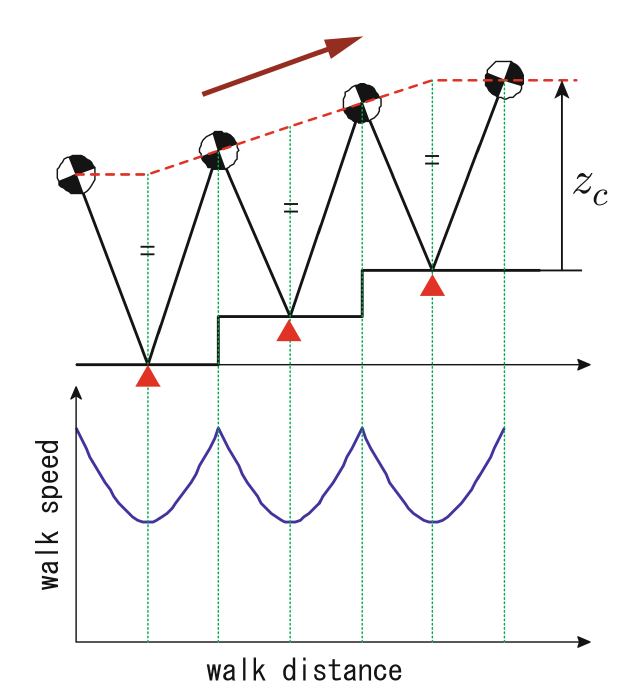}
    \centering
    \caption{Linear inverted pendulum used for walking on stairs. The dotted line is the constraint line \cite{intro_book}.}
    \label{fig:2d-stairs}
\end{figure}
In 1991, Kajita et al. suggested that the above equations do not depend on the structure of the leg and only depend on the constraint parameter \textit{y$_c$}. They introduced the LIPM presenting a biped robot walking five steps and coming to standstill. They managed to show that despite not being able to have a biped with mass-less legs, which is what the LIPM suggests, adding ankle torque and having legs of a small mass, control can be achieved in a similar way.

\subsubsection{3D Linear Inverted Pendulum}
In 1997, Hara, Yokokawa and Sadao extended Kajita et al.'s 2D model to 3D. In this model, we assume the pendulum can rotate about the supporting point freely. The CoM moves on a \textit{constraint plane} this time, which is defined as:
\begin{equation}
    \label{plane}
    z = k_x x + k_y y + z_c
\end{equation}

where \textit{k$_x$}, \textit{k$_y$} determine the slope and \textit{z$_c$} is the height of the constraint plane.

In order for the CoM to move on the plane, its acceleration must be orthogonal to the normal vector to the plane. Using the second derivatives of Equation \ref{plane}
\begin{equation}
    \ddot{z} = k_x \ddot{x} + k_y \ddot{y}
\end{equation}
we can realise the horizontal dynamics of the CoM
\begin{equation}
    \label{y_lipm}
    \ddot{y} = \frac{g}{z_c}y - \frac{1}{mz_c} u_x
\end{equation}
\begin{equation}
    \label{x_lipm}
    \ddot{x} = \frac{g}{z_c}x + \frac{1}{mz_c} u_y
\end{equation}

Kajita et al., presented these equations in 2001 \cite{DBLP:conf/iros/KajitaKKYH01}, where u$_x$ and u$_y$ are virtual input torques around x-axis and y-axis respectively to compensate input non linearity. These equations were later on explained to be used with zero input torques (u$_x$ = u$_y$ = 0).

The horizontal motion of the CoM is not affected by the inclination of the plane, but only affected by z$_c$, the intersection of the plane. 

\begin{figure}[ht]
    \includegraphics[width=5cm,height=4cm]{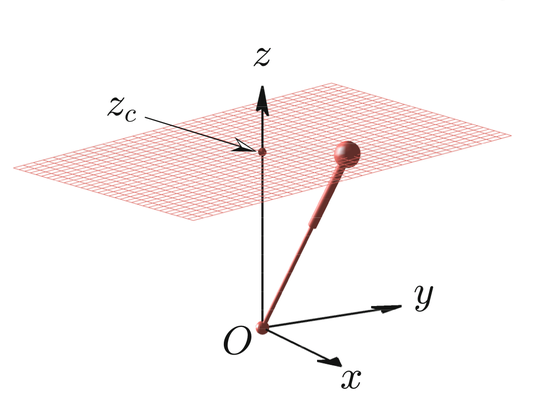}
    \centering
    \caption{The CoM of the 3D linear inverted pendulum moves on the constraint plane \cite{intro_book}.}
    \label{fig:3d-lipm-plane}
\end{figure}

The 3D linear inverted pendulum can be treated as the concatenation of two 2D linear inverted pendulums. Due to the fact we can neglect the inclination of the plane, we can project the trajectories of the two pendulums onto the xy-plane.

It can be observed that transforming to a new reference frame, due to the rotation of the pendulum, the equations of the dynamics are still valid. 
In order to calculate the geometry of the trajectory, orbital energy must be realised. The equations below show the orbital energy of a new frame
\begin{equation}
    \label{orbital_energy1}
    E'_x = -\frac{g}{2z_c} (cx + sy)^2 + \frac{1}{2} (c \dot{x} + s \dot{y})^2
\end{equation}
\begin{equation}
    \label{orbital_energy2}
    E'_y = -\frac{g}{2z_c} (-sx + cy)^2 + \frac{1}{2} (-s \dot{x} + c \dot{y})^2
\end{equation}
where c = $\cos \theta$, s = $\sin\theta,$. By taking the derivatives of the Equations \ref{orbital_energy1} and \ref{orbital_energy2} and setting $\theta = 0$, a hyperbolic equation is obtained for the trajectory of the 3D linear inverted pendulum
\begin{equation}
    \label{hyperbolic}
    \frac{g}{2z_c E_x}x^2 + \frac{g}{2z_c E_y}y^2 + 1 = 0. 
\end{equation}
The hyperbolic trajectory was described by Kajita et al. as an interesting find for biped locomotion, as it applied in similar manner in Keppler motion and the swing-by flight trajectory of spacecraft Voyager 1.

In 2001, Kajita et al. tested the 3D-LIPM with a simple walking pattern generator in a simulation. They managed to make their simulated 12 d.o.f bipedal robot walk along a circle. 

\subsection{3D LIPM and ZMP}
ZMP is widely used in robot research, and control of the ZMP was studied by Kajita et al. along with the 3D LIPM \cite{DBLP:conf/icra/KajitaKKFHYH03}. ZMP can be easily calculated for the 3D LIPM considering the horizontal constraint that $k_x = k_y = 0$ from Equation \ref{plane}
\begin{equation}
    \label{zmp_coord}
    p_x = - \frac{u_y}{mg},
    p_y = \frac{u_x}{mg}
\end{equation}
where ($p_x, p_y$) are the coordinates of the ZMP on the floor. Using the derivatives from Equations \ref{y_lipm} and \ref{x_lipm} and the coordinate Equations \ref{zmp_coord}, we get
\begin{equation}
    \label{y_zmp}
    \ddot{y} = \frac{g}{z_c} (y - p_y)
\end{equation}
\begin{equation}
    \label{x_zmp}
    \ddot{x} = \frac{g}{z_c} (x - p_x)
\end{equation}
In order to output the ZMP coordinates from the model, the Equations \ref{y_zmp} and \ref{x_zmp} were re-arranged to get
\begin{equation}
    \label{py}
     p_y = y - \frac{z_c}{g}\ddot{y}
\end{equation}
\begin{equation}
    \label{px}
     p_x = x - \frac{z_c}{g}\ddot{x}
\end{equation}
These lead to control of the ZMP and Kajita et al. used these with the cart-table model \cite{intro_book} for walking generation based on given ZMP.

\begin{figure}[ht]
    \includegraphics[width=7cm,height=4cm]{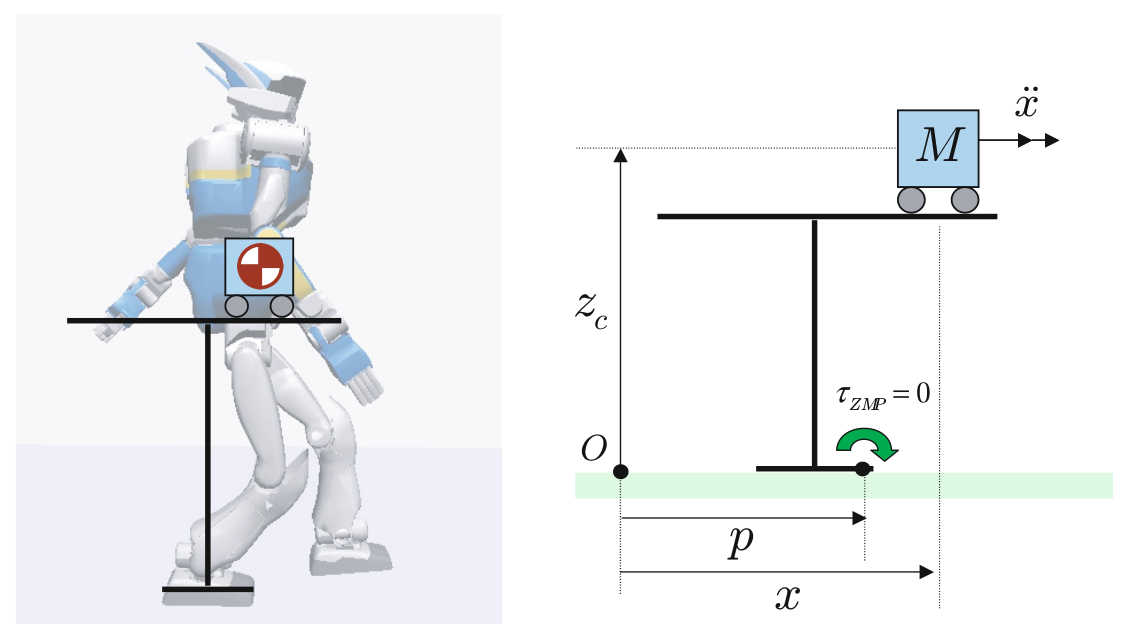}
    \centering
    \caption{Representation of Cart-table model \cite{intro_book}.}
    \label{fig:cart_table_model}
\end{figure}

The Linear Inverted Pendulum is a simple model that can be used for walking pattern generation. Its assumption though, i.e. mass-less legs, make it a bit unrealistic. Also, the fact that the CoM remains at a constant height throughout locomotion makes it seem a bit distant from human-like walking.

\section{Walking Trajectory Generation}\label{trajectory_generation}
Many methods exist that generate walking trajectory and most focus on control of the ZMP. Kajita et al. used control of ZMP using future reference to generate walking pattern \cite{DBLP:conf/icra/KajitaKKFHYH03} \cite{DBLP:conf/iros/KanekoKKYAKOI02}. 

Park et al. presented a novel gait trajectory method \cite{DBLP:conf/icra/ParkKLO06}. Their method is used for online free walking pattern generation and it's mathematically simple with separate trajectory generation on the sagittal and coronal plane. This decoupling of the trajectory on the three axis enhances their method's simplicity. They also managed to generate smooth trajectories due to the use of smooth functions and some proper boundary conditions. Their trajectory generation on the sagittal plane uses trajectory equations for the swing leg ankle position, which is a cycloid function, and the pelvis centre position, which is a three-order polynomial interpolation of the pelvis centre. Their trajectory generation on the coronal plane only has an equation for the pelvis centre in y-frame and it's again a three-order order polynomial interpolation.

In our project, we will use the Polynomial Trajectory generation Algorithm (PTA) \cite{DBLP:journals/ijra/CuevasZPR10}, which is discussed in Section \ref{pta}.

\section{Height Stabilisation}
Controlling the height of the body during locomotion is one of the three controllers which we will discuss later on, in Section \ref{height_control}. Looking at intuitive controls and ways to combine the LIPM and natural locomotion, extensive research was done on Jerry Pratt et al.'s work \cite{DBLP:conf/icra/PrattP98}. They suggested maintaining a constant height during locomotion, which is a perfect fit for our model (LIPM). Their method was to use a virtual spring-damper mechanism, which would cause the body to oscillate about the desired constant height above the ground.

You et al. in a paper trying to implement a biped from Raibert's one legged hopper \cite{DBLP:conf/iros/YouLCT15}, implemented the height controller by controlling the knee joint. They used a three-order polynomial for the trajectory of the knee. With slight modifications to the trajectory planner, they managed to get their biped to walk and run.

We will try to use Inverse Kinematics analysis to handle the height control and the calculations are shown in Section \ref{section:IK}.

\section{Natural Dynamic Mechanisms}
The study of natural locomotion of animals brought forth some features that animals have evolved in order to stabilise their walking trajectory with minimal power and simple control. Jerry Pratt, in his paper \cite{DBLP:phd/ndltd/Pratt00}, presented three natural mechanisms that animals exploit and used them to control Spring Flamingo, the bipedal robot developed for the purpose of researching further into natural dynamics and control. These mechanisms are explained below and their use in this project is examined in order to make walking more natural. They were also combined with the three-part intuitive control suggested by Raibert \cite{Raibert_book}, i.e. controlling height, balance and speed, on the Spring Flamingo \cite{DBLP:conf/icra/PrattP98}, which we will also use in our project. 

\subsection{Knee Cap}\label{knee_cap}
A knee without a mechanism to restrict its motion is unstable. In the case of the support leg during walking, being straight, controlling the knee joint without adding limits will give unwanted results as the knee might bend in the wrong direction. A knee cap is an actual limit stop that prevents the knee from inverting. The technique Jerry Pratt \cite{DBLP:phd/ndltd/Pratt00} suggested was to apply a constant torque, pushing the knee against the cap forcing it to lock when the centre of mass passes the knee cap. This requires no feedback control. 

\begin{figure}[ht]
    \includegraphics[width=8cm,height=4cm]{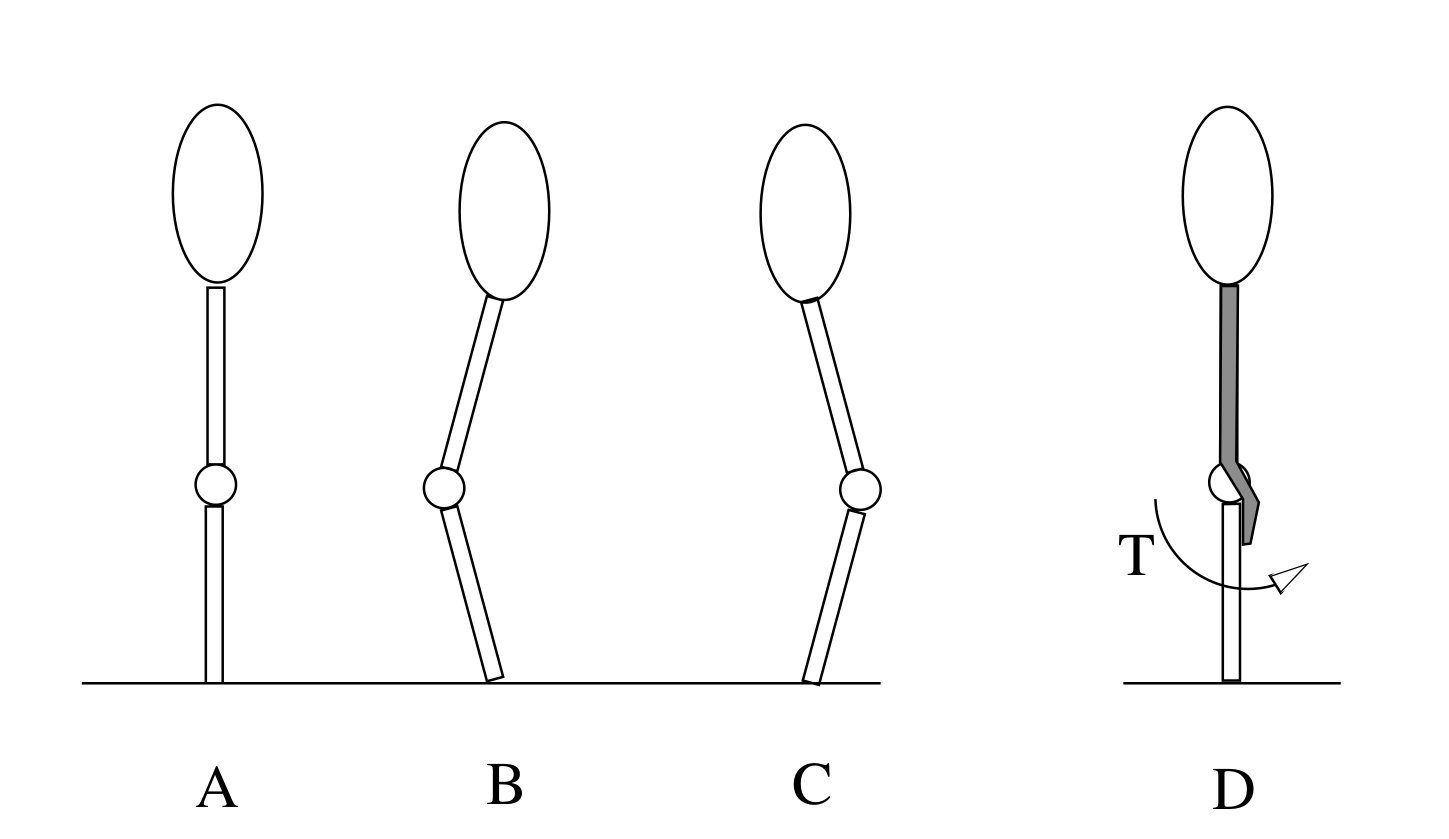}
    \centering
    \caption{A indicates the unstable straight support leg. B and C show the bending that can occur without limits. D shows the knee cap that can lock the knee, stopping it from inverting \cite{DBLP:phd/ndltd/Pratt00}.}
    \label{fig:kneecap}
\end{figure}

\subsection{Compliant ankle}
Ankles help control the speed of motion by applying torque during toe off. More importantly, they smooth the transitions of the centre of pressure on the foot as the centre of mass moves forward. Torque at the ankle can be controlled, but the requirements might be too high due to the centre of pressure being near the toe. Introducing a compliant ankle alongside an actuator for fine control can achieve walking without high torque requirements. Jerry Pratt \cite{DBLP:phd/ndltd/Pratt00} used a quadratic spring and tuned the stiffness parameter during experimentation in a simulated environment. On the Spring Flamingo they used a rubber stop.
\begin{figure}[ht]
    \includegraphics[width=8cm,height=4cm]{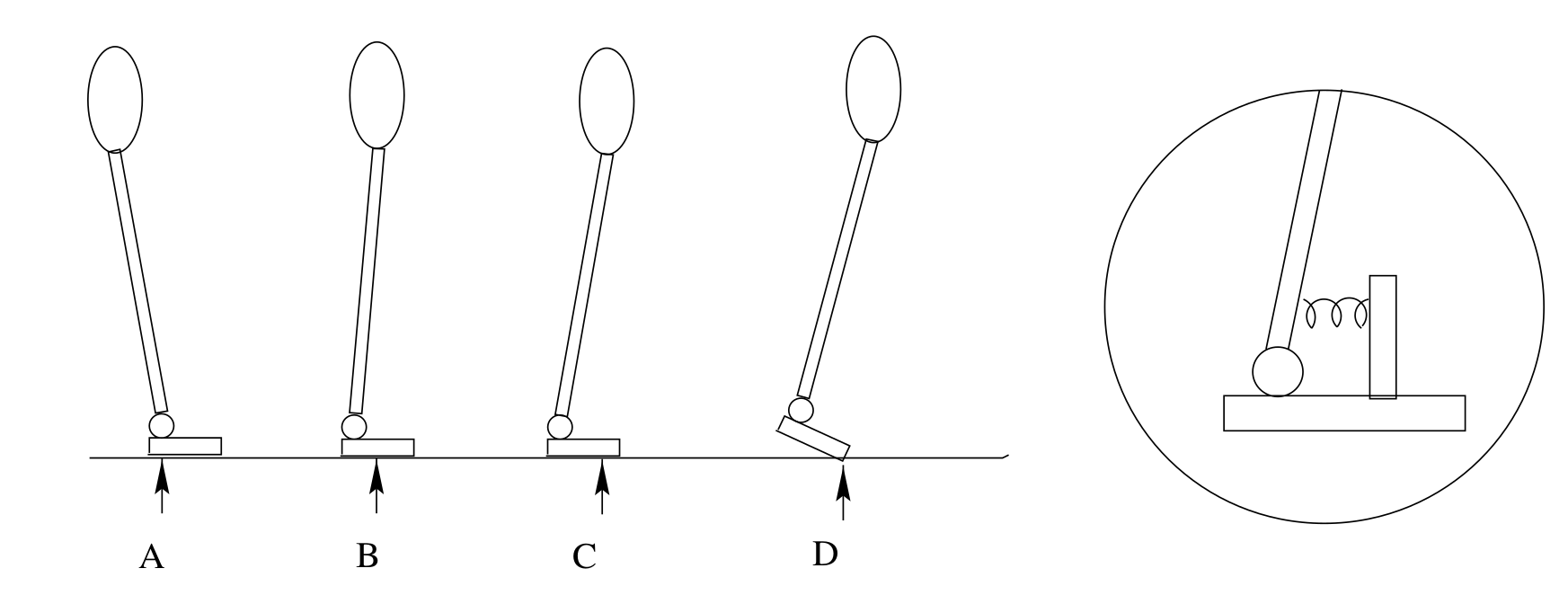}
    \centering
    \caption{ The different stages of the action of centre of pressure as the centre of mass moves forward. The compliant ankle configuration on the right can achieve these transitions without actuation \cite{DBLP:phd/ndltd/Pratt00}.}
    \label{fig:ankle}
\end{figure}
\subsection{Passive swing leg}
Considering passive dynamic walking, the ideal scenario is for a swinging leg to continue its swing until the foot touches the ground again without interventions. The technique here is quite simple again. Hip torque is applied in order to reach a certain angle, while the knee swings freely. In the case of the Spring Flamingo \cite{DBLP:phd/ndltd/Pratt00}, they added damping on the knee joint so that the knee cap is not damaged by the swing, and locked the knee when it hit the cap. This is repeated while exchanging legs illustrating passive swinging.

\section{PID Controller}\label{PID_section}
A Proportional Integral Derivative Controller (PID Controller) is the most popular control loop feedback mechanism used in industry because of its robustness and simplicity. The controller calculates the error term during operation. The error term is a simple calculation as shown below: 
\begin{equation}
    \label{error}
    e(t) = q_d - q_t
\end{equation}
where $q_d$ is the desired value and $q_t$ is the actual measured value.

The controller output $u(t)$ is the sum of the proportional, integral and derivative terms
\begin{equation}
    \label{pid_equation}
    u(t) = K_p e(t) + K_i \int_{0}^{t} e(x) dx + K_d \frac{de(t)}{dt}
\end{equation}
where $K_p$, $K_i$, $K_d$ are the proportional, integral and derivative gains respectively. The controller relies solely on the controlled variable and the output is greatly affected by the gains, which are tuning parameters. Depending on the system requirements, the parameters must be tuned to respond to errors, reduce overshooting from a desired setpoint and regulate oscillations. A natural trade-off exists between speed of response and stability of the system. A fast system results in many oscillations and instability, but on the other hand a very stable system is not fast enough to respond to changes. Another thing to take into consideration is any external disturbances that can occur that will affect the performance of the controller. There exist a few methods to tune the gain parameters, which still happen to have limitations and suffer from this trade-off. Before outlining a few of these methods, let's explain how each of the \textit{P}, \textit{I}, \textit{D} terms affect the controller and its performance. 

\subsection{Proportional}
The \textit{P} Control is simple and powerful, as it provides an output proportional to the error of the system. A large $K_p$ may lead to oscillations around the goal state of the system. The system may also suffer from a steady-state error, where the system reaches the desired setpoint, but we will have an output of zero as the controller operates on error. This can be solved by introducing the Integral.

\subsection{Integral}
Using this, we can estimate the error during operation and as the name suggests we do that by integrating it over time. The Integral is precise and delicate, accumulating the error leading to the goal state. A problem with this is that it may accumulate error unnecessarily, causing the controller to overshoot or oscillate about the goal state. To dampen the oscillations, the Derivative is introduced.

\subsection{Derivative}
The Derivative tries to reduce the rate of change of the error to zero, dampening the oscillations, thus making the system more stable. Usually, the D-term is implemented as $-K_d \dot{q}$, so that $q_d$ is not used, to avoid having a large derivative term, as $\dot{q_d}$ can be a large number due to fluctuations in $q_d$ \cite{derivative}. A general problem with the D-term is it reduces the speed of the control action and it's also sensitive to noise. 

\subsection{Methods for tuning gains}
PID tuning refers to adjusting the three gains to some values that will result in a fast and stable system. This can be quite a difficult task and we will briefly outline a few tuning methods below.

\subsubsection{Manual Tuning}
This is a heuristic method and it can be time consuming. It requires the person implementing the controller to have prior experience in tuning PID parameters for faster and better results. The starting point to this method is to set $K_i$ and $K_d$ to zero and start increasing $K_p$ to the point the output of the control loop oscillates about the goal state. $K_p$ is then set to approximately half of that value. $K_i$ is then increased until the output is corrected. If needed, $K_d$ can be increased to damp any oscillations. Table \ref{pid_table_cited} indicates the effects of the separate gains tuning to the output.

\begin{table}[!ht]
    \centering
    \caption{Effects of independent P, I, and D tuning \cite{pid_table}}
    \begin{adjustbox}{width=1\textwidth}
        \begin{tabular}{|c|c|c|c|c|c|}
            \hline
            Parameter &	Rise time &	Overshoot &	Settling time &	Steady-state error & Stability\\
            \hline
            $K_p$ &	Decrease &	Increase &	Small change &	Decrease &	Degrade\\
            \hline
            $K_i$ &	Decrease &	Increase &	Increase &	Eliminate &	Degrade\\
            \hline
            $K_d$  & Minor change &	Decrease &	Decrease &	No effect in theory	 & Improve if $K_d$ small\\
            \hline
        \end{tabular}
    \end{adjustbox}
    \label{pid_table_cited}
\end{table}

\subsubsection{Ziegler–Nichols method}
Another heuristic method for PID tuning is the Ziegler-Nichols(ZN) method \cite{zn_official_paper}. Again, $K_i$ and $K_d$ are set to zero and $K_p$ is increased until the ultimate gain $K_u$ is reached, having stable oscillations about the goal state. As Table \ref{zn_table} shows, $K_u$ and oscillation period $T_u$ are used for the tuning. The ZN method aims for quarter-amplitude damping response and is suitable for disturbance rejection, as it yields aggressive gain. This results in overshoot which is not ideal for all applications.

\begin{table}[!ht]
    \centering
    \caption{Use of $K_u$ and $T_u$ for ZN tuning method for different controller types \cite{DBLP:journals/tcst/McCormackG98}}
    \begin{minipage}{\textwidth}
        \centering
        \begin{tabular}{|c|c|c|c|}
            \hline
            \textbf{Control Type}    &  \textbf{$K_p$}      &  \textbf{$T_i$}	    &   \textbf{$T_d$}\\
            \hline
            Classic ZN	    &   $0.6K_u$    &   $T_u/2$ 	&   $T_u/8$\\
            \hline
            PIAE            &   $0.7K_u$    &	$T_u/2.5$	&   $3 T_u/20$\\
            \hline
            Some Overshoot  &	$0.33K_u$	&   $T_u/2$	    &   $T_u/3$\\
            \hline
            No Overshoot	&   $0.2K_u$	&   $T_u/2$	    &   $T_u/3$\\
            \hline
        \end{tabular}
    \end{minipage}
    \label{zn_table}
\end{table}

\subsubsection{Cohen-Coon method}
This method aims for quarter-amplitude damping response, similar to the ZN method and also suffers from instability at the occurrence of small changes in the process parameters. Their difference is that the Cohen-Coon(CC) method works on processes where the dead time is less than twice the time constant, whereas the ZN method for dead time less than half the time constant. 

\subsection{Limitations}
PID controllers can be used to solve many control problems and are ideal when there is no model of the process. They have some limitations though making them unable to yield optimal control. This type of control does not have knowledge about the general process but focuses on certain parameters, making it not  robust to factors other than those parameters. 

PID control relies on errors to occur, in order to "fix" them and if tuned poorly will be slow to react. The system might overshoot due to high gains making it unstable, or too slow to react if low gains are used. PID controllers are linear and do not perform well in non-linear systems. The D-term may suffer from noise from the 
measuring devices, resulting to fluctuations in the output.
\newpage
\chapter{Methodology and Modelling}

In this chapter, we will explain our own methodology in approaching the task at hand, bipedal walking. We will use the knowledge and findings of other researchers' work, which was explained in the previous chapter, to implement our own control plan. This chapter will begin by outlining our decomposition of the control task and will proceed to show the controllers we implemented.
\section{Decomposition of Control}
Raibert describes the importance of realising robot legged locomotion as substantial for understanding human and animal locomotion\cite{Raibert_book}. During experimentation, they developed a one-legged hopper robot which they used to simplify the locomotion problem and address the basics of active balance and dynamics. Therefore, they split the control of the robot into three parts, hopping, forward speed and posture control.

\begin{figure}[ht]
    \includegraphics[width=8cm,height=4cm]{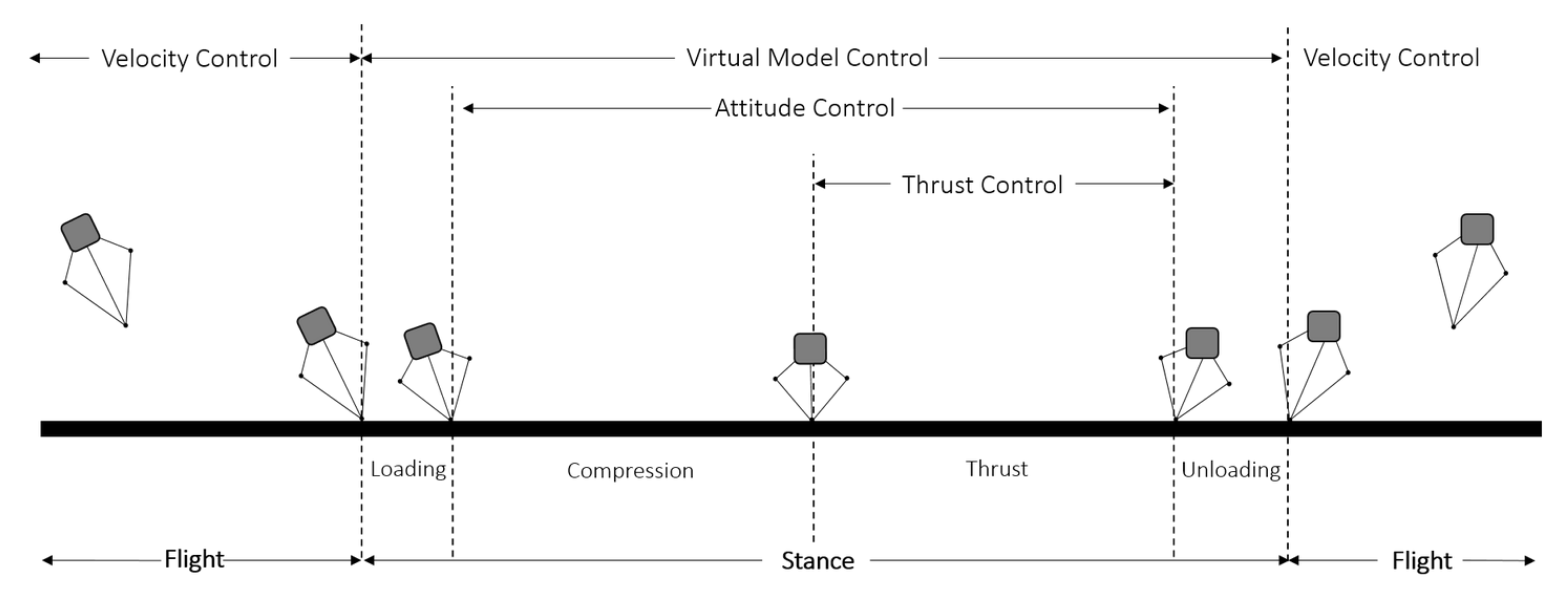}
    \centering
    \caption{Hopper gait phases and control parts \cite{Hopper}.}
    \label{fig:hopper}
\end{figure}
In this project, we are interested in bipedal walking, therefore we will use a different strategy than Raibert's and his colleagues' running/hopping robots. This notion of decomposition of control will be used though in a similar manner, by developing a three-part controller for height, forward speed and balance.

\subsection{LIP Model for a Bipedal Robot}
In order to simplify the walking task and use the above controller, we will use the Linear Inverted Pendulum in 2D, only in a simulated environment. Therefore, we will make use of the information we previously explained in Section \ref{2dLIPMsection}.

\begin{figure}[ht]
    \includegraphics[width=9cm,height=4cm]{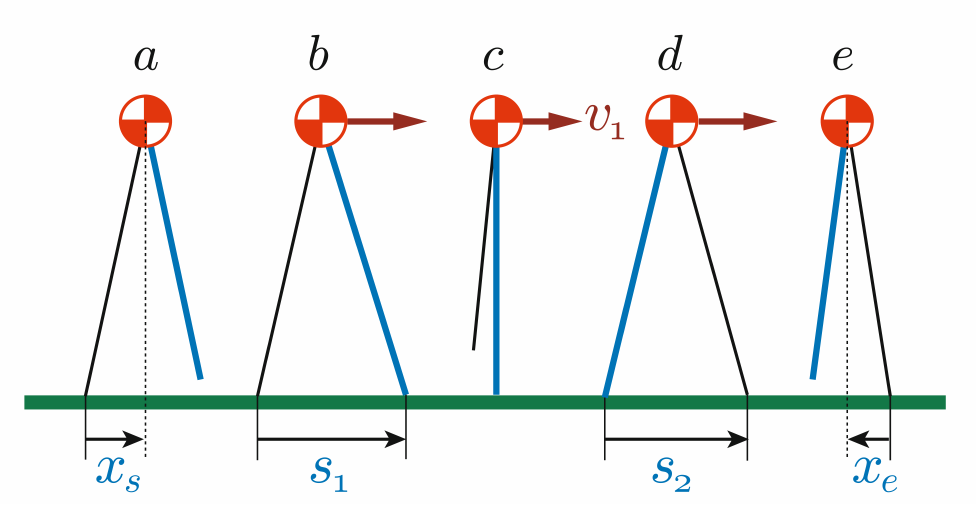}
    \centering
    \caption{Support leg exchange keeping constant height \cite{intro_book}.}
    \label{fig:supportLegExchange}
\end{figure}

\begin{figure}[ht]
    \includegraphics[width=\textwidth,height=4cm]{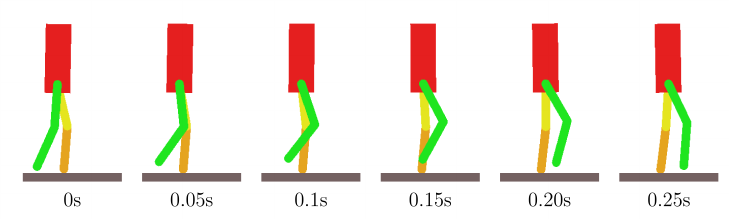}
    \centering
    \caption{Single-support bipedal walking \cite{DBLP:conf/iros/YouLCT15}.}
    \label{fig:singlesupport}
\end{figure}

The whole mass of the robot will be concentrated at its CoM, so the robot will be treated as a point mass with two mass-less legs. Its motion must be linear and the CoM must have constant height during walking. The support leg exchange mechanism used is quite simple, but during walking the robot is always in single-support phase (see Figure \ref{fig:singlesupport}). This means, once the robot starts walking by swinging one of the legs, the other leg is the stance leg and once the swing leg touches the ground, it becomes the stance leg and the other one starts swinging. We can use the LIP model dynamic equations to realise the motion of the bipedal robot, assuming that the swinging legs and the ground contact do not affect the system motion.

\begin{equation}
    \label{x_t}
    x(t) = x(0) \cosh(\frac{t}{T_c}) + T_c\dot{x}(0) \sinh(\frac{t}{T_c})
\end{equation}    
\begin{equation}
    \label{x_t_speed}
    \dot{x}(t) = \frac{x(0)}{T_c} \sinh(\frac{t}{T_c}) + \dot{x}(0) \cosh(t/T_c)
\end{equation}    
\begin{equation}    
    T_c = \sqrt{\frac{h_c}{g}}
\end{equation}

The above equations ensure the horizontal motion of the CoM and its movement in constant height \textit{h}$_c$, because the vertical components of the weight of the point mass are cancelled by the kick force mentioned in Section \ref{2dLIPMsection}. Horizontal displacement $x(t)$ and forward speed $\dot{x}(t)$ of the CoM are determined by the initial conditions $x(0)$ and $\dot{x}(0)$, which are updated after each step.

\subsection{Height Control}\label{height_control}
The first part of the controller regulates the height of the CoM. In our case, the biped controls its height by the joint angles. We do Inverse Kinematics analysis, discussed in Section \ref{section:IK}, to solve the joint angles and we do that always using the constant height specified by the LIPM $h_c$. This analysis, considering feedback received from the joints, ensures the CoM is at the desired $h_c$ at all times. 

\subsection{Forward Speed Control}
\subsubsection{Foot Placement}
The second part of the controller regulates the forward speed and acceleration of the robot. This is done by adjusting the foot placement, i.e. moving the leg from its stance position to another forward position during swing phase. The position of the foot w.r.t the torso of the robot affects its acceleration and speed. In order to find an appropriate forward position for the foot, the actual forward speed, the desired speed and the system's dynamics model need to be considered. In our system, the dynamics model is the LIPM.

The simplicity of Raibert's method of using foot placement to control the forward speed of a robot was used by others to successfully control bipeds and quadrupeds, like the galloping quadruped of Marhefka et al. \cite{quadruped_gallop}. 

\begin{figure}[ht]
    \includegraphics[width=0.8\textwidth,height=6cm]{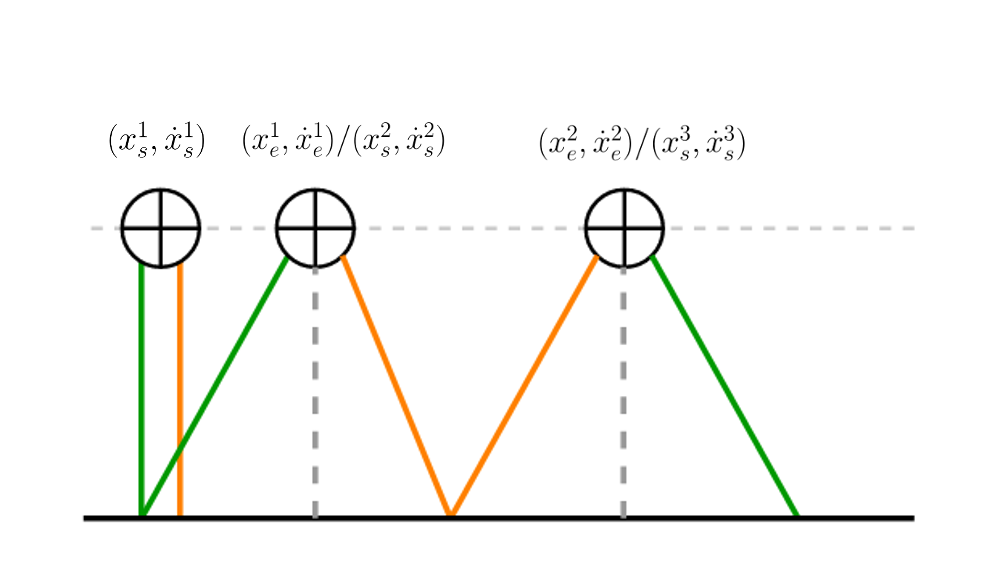}
    \centering
    \caption{State of CoM during a stride.}
    \label{fig:steps_com}
\end{figure}

In order to define foot placement, we need to have a global coordinate system. We will use Raibert's settings and refer to foot placement $p$ of the swing leg w.r.t the position of the CoM. Therefore, in every step the swing leg's foot starts with a negative $p$ and becomes positive when it goes forward surpassing the CoM position. According to the LIPM, we will use the assumption of the ideal scenario, where there is no velocity loss before and after ground contact of the feet. As such, the final velocity of a step is the initial velocity of the next step. 
\begin{equation}
    \label{init_final_vel}
    \begin{array}{lr}
        \dot{x}^{n+1}_s = \dot{x}^n_e, &  n = step \ number\\
    \end{array}
\end{equation}

Therefore, extending from that, we can define $p$ as:
\begin{equation}
    \begin{array}{lr}
        p_n = -x^{n+1}_s, & n = step \ number\\
    \end{array}
\end{equation}

The foot placement can then be expressed as 
\begin{equation}
    \begin{array}{lr}
    p_n = T_c\dot{x}^{n+1}_s \coth{(\tau_s)} - T_c\dot{x}^{n+1}_e \csch{(\tau_s)}, & n = step \ number\\
    \end{array}
    \end{equation}
where $\tau_s = T_{step}/T_c$ is the normalised step time.  We must provide the system with a target velocity $\dot{x}^{n+1}_e$, for its next step. Also, as we are looking ahead in the future, we don't know the value of $\dot{x}^{n+1}_s$. As we are dealing with the LIPM in the ideal scenario, we don't take into account errors between the estimated velocity and the real velocity and fully rely on the Equation \ref{init_final_vel}. Therefore, we can use the LIPM equations to calculate it as
\begin{equation}
    \dot{x}^{n+1}_s = \dot{x}^n_e = \frac{x^n(0)}{T_c} \sinh(\tau_s) + \dot{x}^n(0) \cosh(\tau_s)
\end{equation}  
Using fixed values for $T_c$ and $T_{step}$, a linear expression for the foot placement is obtained. 

Due to experimentation, it could be seen that in reality the performance of the model degrades and work in the field has been done in order to improve this methodology, by calculating some estimate error and considering that, when calculating foot placement \cite{DBLP:conf/humanoids/LiCYVL17}.

In order to improve this method, Raibert proposed tabular control \cite{DBLP:journals/tsmc/RaibertW84}, where a large table of pre-computed data was used describing the position of the feet during locomotion. This was done because of the very similar cycle of the locomotion. Its purpose is to provide predictions about future foot placements, based on a set of "stereotype mannered" variables and a set of "freely varied" variables. A disadvantage of this method is that it requires a lot of experimental data in order to fill the table with the sets and they need to be well approximated. It also operates offline, therefore does not respond quickly to disturbances in the environment.

You et al. proposed another improved method, using online linear regression analysis for the foot placement control \cite{DBLP:conf/iros/YouLCT15}. The method does not use a lot of data and also works online, enabling quick response to disturbances. It was used on a one-legged hopper for comparison with Raibert's method and performed better, producing stable hopping. They extended the controller to a bipedal robot, producing stable walking and running (human jogging speed). In both cases accurate forward velocity tracking was achieved.

\subsubsection{Foot Trajectory}\label{pta}
In order to realise foot placement, we need gait trajectory as well. Some methods were discussed in Section \ref{trajectory_generation}, but we will use the Polynomial Trajectory Algorithm (PTA) \cite{DBLP:journals/ijra/CuevasZPR10}, because of the smooth trajectories produced and its simplicity. PTA is based on cubic Hermitian polynomial interpolation \cite{Hermit_encyclopedia} of the kinematics of the robot and it can produce trajectories in sagittal plane (hip and swing leg foot) and hip trajectory in frontal plane.

In our case, we need foot trajectory for the swing leg in the sagittal plane and PTA provides equations for the foot position, i.e. in both x and z-axis as $x_f(t)$ and $z_f(t)$ respectively
\begin{equation}
    \begin{array}{cc}
        x_f(t) = x_{fs} + 3(x_{fe} - x_{fs})\frac{(t - kT)^2}{T_s^2} - 2(x_{fe} - x_{fs})\frac{(t - kT)^3}{T_s^3} &  0 < t \leq T_{step}
    \end{array}
\end{equation}
\begin{equation}
    z_f(t) = \begin{cases}
        z_{fs} + 3(z_{fm} - z_{fs})\frac{(t - kT)^2}{T_m^2} - 2(z_{fm} - z_{fs})\frac{(t - kT)^3}{T_m^3} &  0 < t \leq T_m \\
        z_{fm} + 3(z_{fe} - z_{fm})\frac{(t - kT - T_m)^2}{(T_s - T_m)^2} - 2(z_{fe} - z_{fm})\frac{(t - kT - T_m)^3}{(T_s - T_m)^3} &  T_m < t \leq T_{step}
    \end{cases}
\end{equation}
where $s$ and $e$ indicate the start and end of a step, $z_{fm}$ is the maximum height reached by the foot and $T_m$ is the time it take for the foot to reach $z_{fm}$. 

It's important that the parameters $T_m$ and $z_{fm}$ mentioned above are chosen carefully, as they will greatly affect the velocity and acceleration of the robot, thus affecting the smoothness of the trajectory. In order to have smooth trajectory, the contact with the ground must be smooth and this is achieved if the swing leg foot's speed at contact is close to zero. The ideal parameter setting is $T_m = T_{step}/2$ for smooth impact transition.

\subsection{Balance Control}\label{balance_control}
The third part of the controller regulates the balance of the robot. When discussing about balance, we are referring to the torso attitude. The body is desired to be kept at zero pitch angle. Looking at Li et al. papers \cite{DBLP:conf/iros/LiZZXTC15} \cite{DBLP:journals/trob/LiZZX17}, body posture is controlled by adding a "compensating" torque at the hips. This torque is decided by controlling the rate of change of the body angle.

\begin{equation}
    \begin{cases}
        \omega_d(i) = K_p(\phi_{ref}(i) - \phi_t(i)) - K_d\dot{\phi}_t(i),\\
        \phi_d(i) = \phi_d(i-1) + \omega_d(i)T,
    \end{cases}
\end{equation}
where $\phi_t$ is the measured body angle, $\dot{\phi}_t$ is the measured angular velocity of body and $K_p$ and $K_d$ are the control gains, explained further in Section \ref{upright_pd_section}. The second equation from above is numerical integration of the desired angular velocity $\omega_d$, which updates the body orientation $\phi_d$.

\section{Inverse Kinematics: Calculate joint angles}\label{section:IK}
In order to have a more realistic and natural locomotion, Inverse Kinematics is used to calculate the joint rotation for the different rotary parts of a robot. Our simulated robot will have hip, knee and ankle joints, therefore we need to split our telescopic legs (pendulum form) into upper legs(thighs)and lower legs(shins), linked by the knee joints and feet linked to the shins by the ankle joints. For simplicity, thighs and shins are assumed to be of equal length $L$. Each leg has a related hip, knee and ankle angle denoted as $(\gamma_i, \theta_i, \xi_i)$ for $i = 1,2$ indicating the leg. Hip angles $\gamma$ are defined w.r.t the CoM, whereas knee angles $\theta$ are defined w.r.t its thigh. Ankle angles $\xi_i$ are defined w.r.t its shin.

We mentioned the split of telescopic legs, which we will call virtual legs from now on. In the following figures, the virtual legs $L_{v1}$, $L_{v2}$ form isosceles triangles with the thighs and shins of length $L$. As we are working in 2-D space, the calculations are simple trigonometry using sine and cosine rules.

\begin{figure}[ht]
    \centering
    \begin{minipage}{0.5\textwidth}
	    \centering
	    \includegraphics[width=6cm, height=7cm]{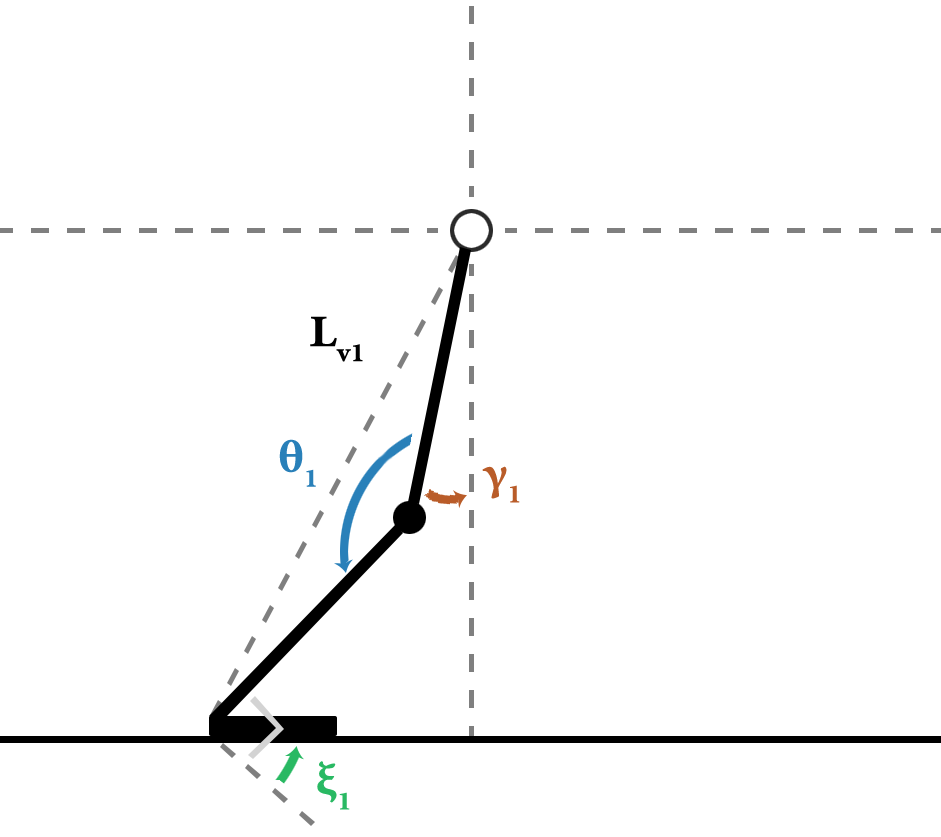}
	    \subcaption{Overview of support leg.}
    \end{minipage}%
    \begin{minipage}{0.5\textwidth}
	    \centering
	    \includegraphics[width=6cm, height=7cm]{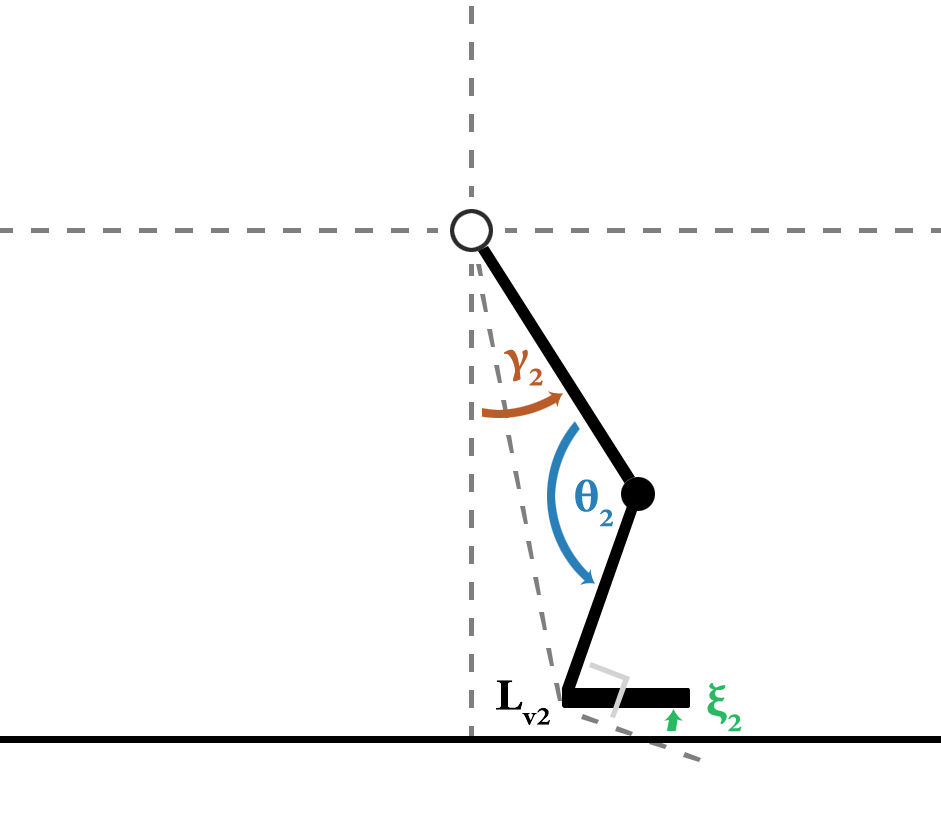}
	    \subcaption{Overview of swing leg.}
    \end{minipage}%
    \centering
    \caption{Inverse Kinematics analysis}
    \label{ik_analysis}
\end{figure}

The Inverse Kinematics analysis is split into calculations for the support leg and the swing leg, in order to find out the hip and knee angles. The support leg calculations are simpler and are the following:
\begin{equation}
    L_{v1} = \sqrt{x_t^2 + (h_c - h_f)^2}
\end{equation}
\begin{equation}
    \theta_1 = \arccos(\frac{2L^2 - L_{v1}^2}{2L^2})
\end{equation}
\begin{equation}
    \gamma_1 = \frac{\pi - \theta_1}{2} - \arctan(\frac{x_t}{h_c - h_f})
\end{equation}
\begin{equation}
    \xi_1 = \frac{\pi}{2} - (\frac{\pi}{2} - (\pi - \theta_1) - \gamma_1)
\end{equation}
where $x_t$ is the x-coordinate of the CoM, $h_c$ is the constant height we assume due to the LIPM and $h_f$ is the height of the foot.

The swing leg calculations will need to consider the position of the foot ($x_{ft}, z_{ft}$)
\begin{equation}
    L_{v2} = \sqrt{(x_{ft} - x_t)^2 + (h_c - z_{ft})^2}
\end{equation}
\begin{equation}
    \theta_2 = \arccos(\frac{2L^2 - L_{v2}^2}{2L^2})
\end{equation}
\begin{equation}
    \gamma_2 = \frac{\pi - \theta_2}{2} - \arctan(\frac{x_{ft} - x_t}{h_c - z_{ft}})
\end{equation}
\begin{equation}
    \xi_2 = \frac{\pi}{2} - (\pi - (\frac{\pi-\theta_2}{2} + (\frac{\pi}{2} - \arcsin{\frac{h_c - z_{ft} - h_{ft}}{L_{v2}}})))
\end{equation}

Using Inverse Kinematics we can control the CoM position or foot placement by controlling the joint angles. It's important to ensure these calculated joint angles do not exceed the physical limits of the robot, so that no damages occur. This good practice was adopted in the simulation as well. Virtual limits were added at the knee joints, acting like knee caps mentioned in Section \ref{knee_cap}, stopping the shin from turning at an angle greater than \ang{180} from the thigh.

\section{Simulated Environment}
Everything explained in this chapter is used to implement a control system for biped walking in a simulated environment. The OpenAI Gym toolkit \cite{DBLP:journals/corr/BrockmanCPSSTZ16} contains an existing environment called Box2D, which implements a reinforcement learning algorithm achieving walking of a four-joint biped robot. In this project, we will utilise that existing simulation library, as well as the robot structure and the LIP-based control system implemented by Cheng Z. for his Masters thesis in the same environment \cite{Zeyu_project}, to create a more refined control system. The codebase was optimised and a full three-part controller was implemented.

\subsection{Robot Structure and Parameters}\label{inital_simulation}
\begin{figure}[ht]
    \centering
    \begin{subfigure}[t]{0.33\textwidth}
	    \centering
	    \includegraphics[width=2.5cm, height=4cm]{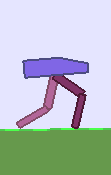}
	    \caption{}
	    \label{robot_rl}
    \end{subfigure}%
    \begin{subfigure}[t]{0.33\textwidth}
	    \centering
	    \includegraphics[width=2.5cm, height=4cm]{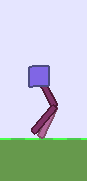}
	    \caption{}
	    \label{robot_zeyu}
    \end{subfigure}
    \begin{subfigure}[t]{0.33\textwidth}
	    \centering
	    \includegraphics[width=2.5cm, height=4cm]{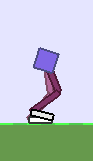}
	    \caption{}
	    \label{robot_mine}
    \end{subfigure}%
    \centering
    \caption{Different structures of robot: (a) Initial Box2D robot structure; (b) Robot structure by previous student \cite{Zeyu_project}; (c) Robot structure used in this project.}
    \label{robot_structures}
\end{figure}

The original Box2D robot consisted of two rectangular legs, split in thighs and shins, and a trapezoid as the body. We will use the modified robot structure from Cheng's Masters thesis \cite{Zeyu_project}, which uses the same legs but a square torso. The two structures are shown in Figure \ref{robot_structures}. The Box2D environment allows the creation of robot parts in the form of Dynamic Body types, which are connected by revolute joints. At first, a four-joint robot was created with four revolute joints which have a single degree of freedom. The hip joints are connected at the centre of the square torso and the thighs and shins are of equal length and are connected by the knee joints. A further addition we did to the robot are feet and two ankle joints, forming a final robot structure of six joints. The dimensions of the different robot parts are shown in Table \ref{robot_params_table}. 

The LIP model suggests the whole mass being at the pendulum, therefore we wanted to pursue the assumption of mass-less legs for the robot. This cannot happen in reality, so the ideal was to keep the mass of the legs as small as possible, at a mass ratio of 10:1 for torso:shin/foot and 8:1 for torso:thigh.

Another important thing to consider is the constant height the CoM needs to be at all times, which was set to be at a factor of 0.9 of the length of the leg (\textit{thigh + shin + foot}). The ground surface was set to have a high friction coefficient, so that the robot feet don't slide, helping it to balance but not affecting the LIP model generating the walking pattern. 

\begin{table}[!ht]
    \centering
    \caption{Robot structure parameters}
    \begin{adjustbox}{width=1\textwidth}
      \begin{tabular}{|c|c|c|c|c|c|}
        \hline
        Body Part  &  Length   &  Width   &  Mass   &  Planar Density  &  Friction Coefficient\\
                   &    ($m$)  &   ($m$)  &  ($kg$) &    ($kg/m^2$)    &\\
        \hline
        Torso  &  0.29  &  0.29  &  0.42  &  5.0  &  0.1\\
        \hline
        Thigh  &  0.57  &  0.09  &  0.05  &  1.0  &  0.1\\
        \hline
        Shin  &  0.57  &  0.07  &  0.04  &  1.0  &  0.1\\
        \hline
        Foot  &  0.38  &  0.1   &  0.038  &  1.0  &  0.1\\
        \hline
      \end{tabular}
    \end{adjustbox}
    \label{robot_params_table}
\end{table}

The initial condition of the system $(x(0), \dot{x}(0))$ is vital for the performance of the robot, as the Equations \ref{x_t} and \ref{x_t_speed} suggest. It cannot be set to zero as the resulting displacement and velocity will result to zero as well, making the robot to stay still. As we don't allow external forces to act on the robot, apart from gravity, the initial condition was set to $(+0.25m, 0 m/s)$. The initial condition we set is further examined in our Results chapter, highlighting its effect on the overall performance of the robot. Gravity then plays its role in giving a natural forward movement to the robot, as the CoM is slightly ahead its support point.   


\subsection{PD Controllers}\label{PD_Controllers_section}
In Box2D, the robot moves based on the torques provided by its joint motors. These torques are decided based on the robot kinematics, i.e. the robot's joint angles taken by the inverse kinematics analysis. This method is called position control and the desired and actual angles can easily be compared and used to compute the error term in a PID controller, as described in Section \ref{PID_section}. 

In this project, we will use PD Controllers, disregarding the Interval term, thus setting it to zero. Due to position control, we can use the joint angle difference from the desired angle as the error term and the angular velocity for the D-term, thus producing this PD controller: 
\begin{equation}
    \label{pd_equation}
    u(t) = K_p e(t) - K_d \dot{q}
\end{equation}
The controller output $u(t)$ is used to compute the torques applied to the robot joints. The simulation has an internal method for doing so. In this project, we implemented our own PD controllers, for each hip, knee and ankle joint. Another PD controller was also added on the hip joints for upright posture control. 

\subsubsection{Hip PD Controller}\label{hip_pd_controller}
The implementation of the hip PD Controller is based on the Equation \ref{pd_equation}:
\begin{equation}
    \label{hip_equation}
    u_{hip} = K_p (\alpha_d - \alpha) - K_d\dot{\alpha}
\end{equation}
where $\alpha_d$ is the desired hip angle, $\alpha$ is the hip angle measured from the virtual sensors and $\dot{\alpha}$ is the angular velocity. 

Two different tuning procedures were followed for the hip joints, that yielded different results and were considered to be used for different purposes. The first one was to set gravity to zero, creating a weightless environment. The robot was "pinned" hanging above the ground and torques were applied to the hip joints with the desired angle being the sine-wave over time. Parameter tuning was done under these circumstances until a satisfactory control performance was achieved.

The second one was to set the robot in standstill, with its feet "pinned" to the ground, the knees fully extended and then applied torque at the hip joints, thus moving the torso. The body would fall because of the torque applied at the hips, therefore we enlarged the shins to form a wide base, so that the robot would not fall. The desired angle used was a sine-wave applied over time. Again, parameter tuning was done under these circumstances until a satisfactory control performance was achieved. 

\begin{figure}[ht]
    \label{fig:hip_tuning}
    \centering
    \begin{minipage}{0.5\textwidth}
	    \centering
	    \includegraphics[width=5cm, height=4cm]{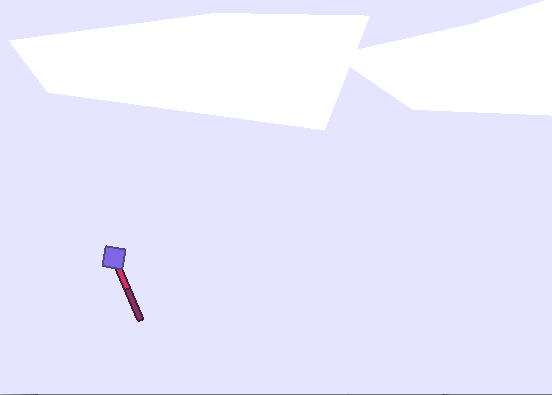}
	    \subcaption{"Pinned-above-ground", no-gravity method.}
    \end{minipage}%
    \begin{minipage}{0.5\textwidth}
	    \centering
	    \includegraphics[width=5cm, height=4cm]{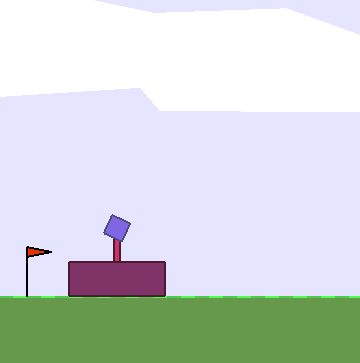}
	    \subcaption{"Pinned-to-the-ground" method.}
    \end{minipage}%
    \centering
    \caption{Robot during hip controller PD tuning.}
    \label{hip_pd_methods}
\end{figure}

One idea was to use the first set of parameters, which were larger, for the support leg as it needs to move the hip and holds the mass of the body as well and the second set for the swing leg where the hip joint only moves the leg and no further masses. 

\subsubsection{Knee PD Controller}\label{knee_pd_controller}
The implementation of knee PD Controller is based on the Equation \ref{pd_equation}:
\begin{equation}
    \label{knee_equation}
    u_{knee} = K_p (\beta_d - \beta) - K_d\dot{\beta}
\end{equation}
where $\beta_d$ is the desired knee angle, $\beta$ is the knee angle measured from the virtual sensors and $\dot{\beta}$ is the angular velocity. 

The tuning procedure followed for the knee joints was to set the robot upside down, keep the thighs at a $\ang{90}$ at all times and apply torques at the knee joints, moving the shins. The robot would lose balance, therefore we enlarged the torso to form a wide base, so that the robot would not fall. Again, the desired angle used was a sine-wave applied over time and parameter tuning was done until a satisfactory control performance was achieved. 
\begin{figure}[ht]
    \includegraphics[width=5cm,height=3cm]{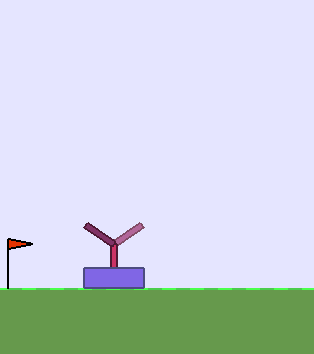}
    \centering
    \caption{Robot during knee controller PD tuning.}
    \label{fig:knee_tuning}
\end{figure}
\subsubsection{Ankle PD Controller}\label{ankle_pd_controller}
The implementation of ankle PD Controller is again based on the Equation \ref{pd_equation}:
\begin{equation}
    \label{ankle_equation}
    u_{ankle} = K_p (\gamma_d - \gamma) - K_d\dot{\gamma}
\end{equation}
where $\gamma_d$ is the desired ankle angle, $\gamma$ is the ankle angle measured from the virtual sensors and $\dot{\gamma}$ is the angular velocity.

The tuning procedure followed for the ankle joints was similar to the knee joint tuning procedure. The robot was set upside down with straight legs (thighs and shins), applied torques to the ankle joints, moving the feet. The desired angle was again a sine-wave applied over time.
\begin{figure}[ht]
    \includegraphics[width=5cm,height=4cm]{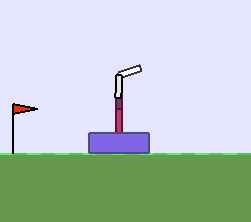}
    \centering
    \caption{Robot during ankle controller PD tuning.}
    \label{fig:ankle_tuning}
\end{figure}

\subsubsection{Upright Posture PD Control}\label{upright_pd_section}
The torques applied at the hips cause the torso to tilt, as the hip joints are connected at the centre of the torso. In order to prevent this tilting, we implement an upright posture controller, as explained in Section \ref{balance_control}, which makes the torso be upright at all times. At the moment, we are working on flat ground, therefore further adjustments about the inclination of the ground are not required. 

The implementation of this PD Controller is again based on the Equation \ref{pd_equation}:
\begin{equation}
    \label{torso_equation}
    \omega^d = K_p (\theta_d - \theta) - K_d\dot{\theta}
\end{equation}
where $\theta_d$ is the desired torso angle, $\theta$ is the torso angle measured from the virtual sensors and $\dot{\theta}$ is the angular velocity.  

This PD Controller is an adjustment controller, as its output will be added to the desired angle fed to the hip PD Controllers. The equation is the following:
\begin{equation}
    \label{angular_desired}
    q_{hip(i)} = q^d_{hip(i-1)} + \omega^d t
\end{equation}
where $q_{hip(i)}$ is the angle provided to the hip PD Controller replacing the desired hip angle, $q^d_{hip(i-1)}$ is the desired angle decided by the kinematics calculations for the hip accumulated with the previous adjustments and $\omega^d$ is the output from Equation \ref{torso_equation}. 

This accumulation of hip angle adjustment is done during the stance phase. In our situation of single support, the accumulation is done throughout walking on the supporting leg. Once the supporting leg, changes phase to swing, the angle is cleared to zero and the accumulation begins on the other leg entering the stance phase. This will eliminate the tilting of the torso.

\subsection{Low-Pass Filter}\label{low_pass_filter}
The knowledge we have on the Box2D environment and the performance shown in the previous work \cite{Zeyu_project}, led us to pay attention in possible high-frequency signal mutations of the measured variables during PD control. Our experiments shown in Section \ref{low_pass_experiments}, confirm the high-frequencies of the signals of the joints actuators. Thus, we implement a low-pass filter on the derivative term of our PD controllers to attenuate these measurements and smooth the signal from the actuators. 

As indicated by \cite{DBLP:conf/icra/QianS92} and used by \cite{derivative} on PID controllers, a low-pass filter with a cut off frequency sufficiently below the system open loop resonance frequency in the forward control path can tackle the instability caused by the system's sampling rate. We implement all PD controllers in this manner:
\begin{equation}
    \label{filter_equation}
    u(t) = K_d (q_d - q) - \frac{a}{s+a}  K_d \dot{q}
\end{equation}
where $a$ is a small value compared to the system open loop resonance frequency $s$. Therefore, we filter the measured velocity before we pass it to the D term of the PD controller. The measured angle was deemed unnecessary to be filtered, because as theory suggests there won't be significant signal distortions for the angle signal, therefore the P term of the PD controller will not be affected. In any case, we experimented with the angle signal as well and confirmed the theory.

\newpage

\chapter{Experimentation and Results}

The previous chapter was a detailed explanation of our use of the LIP model, the Inverse Kinematics and the Controllers necessary for achieving bipedal walking. This chapter will focus on the experimentation done in the simulated environment and will show detailed analysis of the results. 

\section{System Parameters and Model Parameters}
As the equations for LIPM and foot trajectory suggest, some constant parameters had to be chosen, that were vital for the execution of our task. After some experimentation and manual adjustments, Table \ref{params} shows the parameters we use in our calculations.

\begin{table}[!ht]
    \centering
    \caption{System and Model Parameters used during experimentation.}
    \begin{minipage}{0.49\textwidth}
        \subcaption{System Parameters}
        \centering
        \begin{tabular}{|c|c|}
            \hline
            \textbf{Parameter}  & \textbf{Value} \\
            \hline
            Ground Friction     &  2.5\\
            \hline
            $\omega_{hip}$      &  4$rad/s$\\
            \hline
            $\omega_{knee}$     &  6$rad/s$\\
            \hline
            $\omega_{ankle}$    &  4$rad/s$\\
            \hline
            $\tau$              &  100 $Nm$\\ 
            \hline
        \end{tabular}
    \end{minipage}
    \begin{minipage}{0.5\textwidth}
        \subcaption{Model Parameters}
        \centering
        \begin{tabular}{|c|c|}
            \hline
            \textbf{Parameter} & \textbf{Value} \\
            \hline
            $h_c$       &  1.11$m$\\
            \hline
            $T_{step}$  &  0.4$s$\\
            \hline
            $T_m$       &  $T_{step}/2  = 0.2s$\\ 
            \hline
            $z_{fm}$      &  $h_c/5  = 0.222m$\\
            \hline
            $v_d$       &  $0.6m/s$\\
            \hline
        \end{tabular}
    \end{minipage}
    \label{params}
 \end{table}

\begin{figure}[ht]
    \includegraphics[width=8cm,height=6cm]{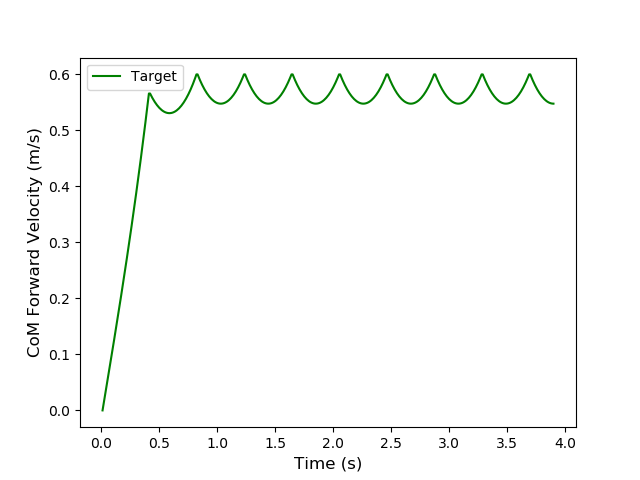}
    \centering
    \caption{Desired forward velocity of CoM over time.}
    \label{fig:desired_com_vel}
\end{figure}

In all our experiments, the desired forward velocity of the CoM, $v_d$, is set at $0.6m/s$. This constant velocity has the potential result of the CoM to travel at a nearly constant velocity. As Figure \ref{fig:desired_com_vel} shows, the CoM will move periodically after the initial acceleration to reach $v_d$. We can verify the LIPM dynamics, as we can see that the maximum velocity the CoM will reach is the $v_d$ and the minimum velocity will be when the body is exactly above the support point.

\section{Experimentation}
This section will show the improvement of the Low-Pass Filter implemented, provide the results of the PD tuning methods we explained in Section \ref{PD_Controllers_section} and then thoroughly explain the different approaches we decided to take, in order to tackle the bipedal walking task. We will show the performance of the biped robot with point feet and three methods where we add feet.
\begin{figure}[ht]
    \centering
    \begin{minipage}[b]{0.5\textwidth}
	    \centering
	    \includegraphics[width=7cm, height=5cm]{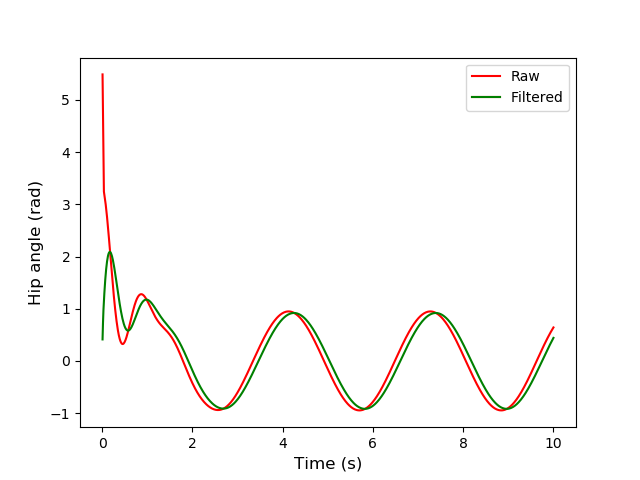}
	    \subcaption{Hip angle.}
    \end{minipage}%
    \begin{minipage}[b]{0.5\textwidth}
	    \centering
	    \includegraphics[width=7cm, height=5cm]{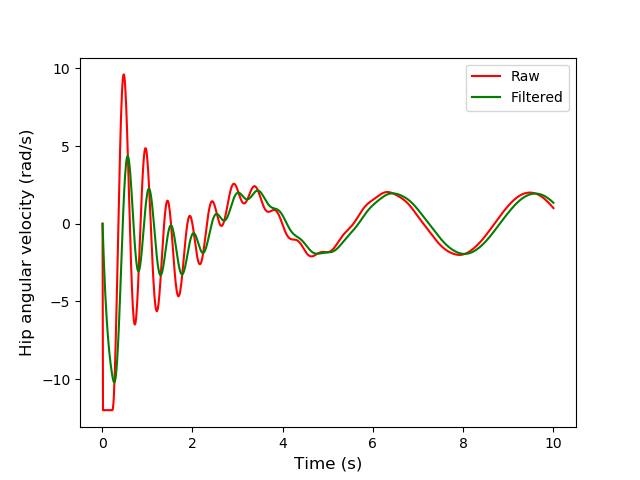}
	    \subcaption{Hip angular velocity.}
    \end{minipage}%
    \centering
    \caption{Comparison of Actual vs Filtered measured variables produced by hip PD Controllers in "Pinned-in-the-air” method.}
    \label{fig:hip_filter_gr}
\end{figure}
\subsection{Low-Pass Filter}\label{low_pass_experiments}
As mentioned in Section \ref{low_pass_filter}, we implemented a low-pass filter on the D term of all our PD controllers. We did experiments during the PD tuning process, to confirm the high-frequency signal mutations of the measured angular velocity. We also did experiments on the measured angle, to ensure that a low-pass filter is not necessary for. 
\begin{figure}[ht]
    \centering
    \begin{minipage}[b]{0.5\textwidth}
	    \centering
	    \includegraphics[width=7cm, height=5cm]{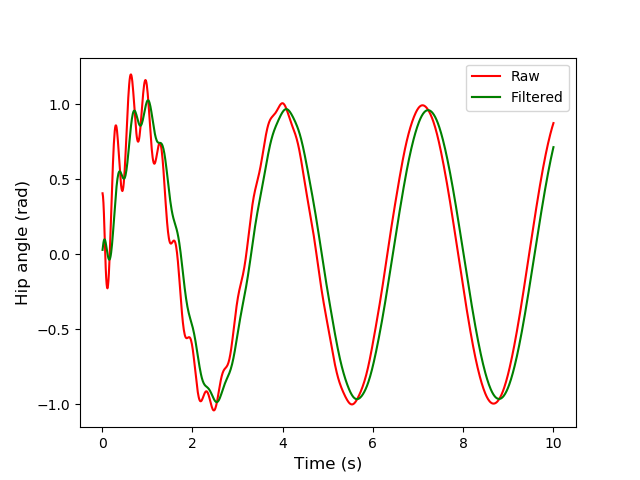}
	    \subcaption{Hip angle.}
    \end{minipage}%
    \begin{minipage}[b]{0.5\textwidth}
	    \centering
	    \includegraphics[width=7cm, height=5cm]{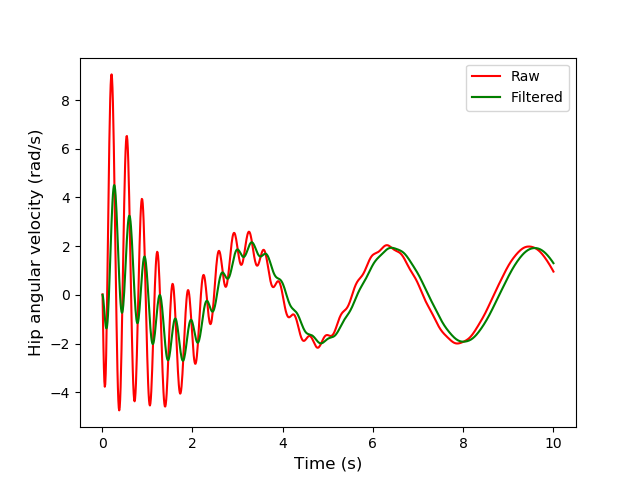}
	    \subcaption{Hip angular velocity.}
    \end{minipage}%
    \centering
    \caption{Comparison of Actual vs Filtered measured variables produced by hip PD Controllers in "Pinned-on-the-ground” method.}
    \label{fig:hip_filter_air}
\end{figure}

\begin{figure}[ht]
    \centering
    \begin{minipage}[b]{0.5\textwidth}
	    \centering
	    \includegraphics[width=7cm, height=5cm]{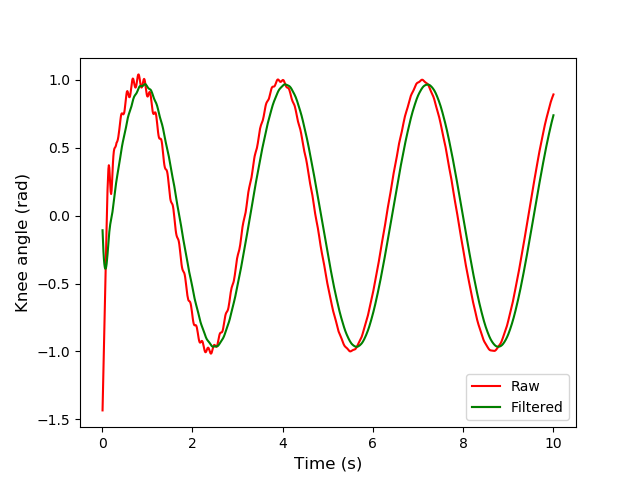}
	    \subcaption{Knee angle.}
    \end{minipage}%
    \begin{minipage}[b]{0.5\textwidth}
	    \centering
	    \includegraphics[width=7cm, height=5cm]{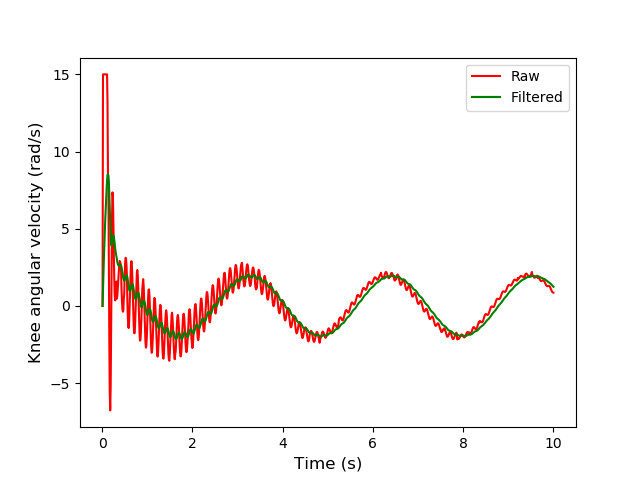}
	    \subcaption{Knee angular velocity.}
    \end{minipage}%
    \centering
    \caption{Comparison of Actual vs Filtered measured variables produced by knee PD Controller.}
    \label{fig:knee_filter}
\end{figure}
\newpage
\begin{figure}[ht]
    \centering
    \includegraphics[width=7cm, height=5cm]{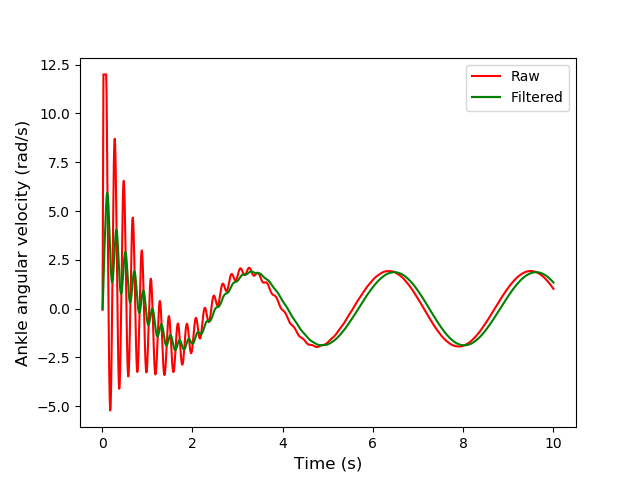}
	\caption{Comparison of Actual vs Filtered angular velocity produced by ankle PD Controller.}
    \label{fig:ankle_filter}
\end{figure}
The experiments were done in the first phase of the robot, therefore we only have figures for the hip and knee angles and angular velocity. Later on, when we proceeded to add feet, we went straight away and only filtered the angular velocity of the ankle joints, without bothering about the angle. Figures \ref{fig:hip_filter_gr}, \ref{fig:hip_filter_air}, \ref{fig:knee_filter}, \ref{fig:ankle_filter} show our experiments on hip (both PD tuning methods), knee and ankle joints respectively. 

The hypothesis about the high-frequency mutations of the measured angular velocity and the insignificance of filtering the measured angle was correct. We also observe the low-pass filter affects the performance of the PD controllers, despite the smoothing of the signals. The $a$ value used was carefully tuned, getting to $a = 0.075$, in order to get the best results. The PD tuning procedure was done with the filtering.
\subsection{PD Tuning}

\subsubsection{Hip PD Tuning Results}
\begin{figure}[ht]
    \centering
    \begin{minipage}[b]{0.5\textwidth}
	    \centering
	    \includegraphics[width=7cm, height=5cm]{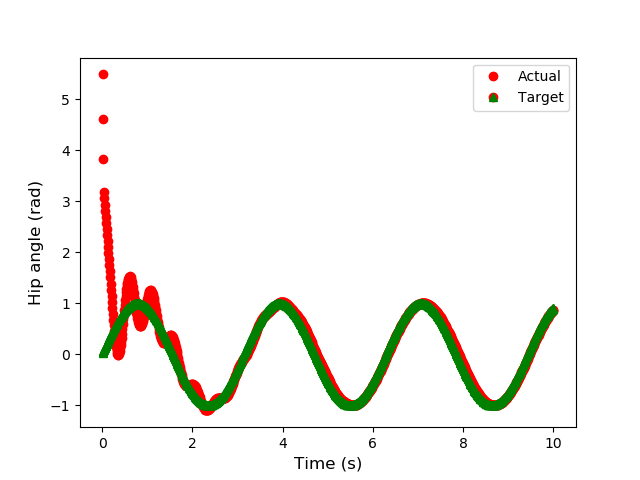}
	    \subcaption{Desired vs Actual Hip Angle over time.}
    \end{minipage}%
    \begin{minipage}[b]{0.5\textwidth}
        \centering
        \begin{tabular}{|c|c|}
            \hline
            \textbf{Parameter} & \textbf{Value} \\
            \hline
            $K_p$       &  100.5\\
            \hline
            $K_d$       &  5.0\\
            \hline
        \end{tabular}
        \subcaption{PD Gains}
    \end{minipage}%
    \centering
    \caption{"Pinned-above-ground" method hip controller performance and gains.}
    \label{fig:hip_pd_air}
\end{figure}

\begin{figure}[ht]
    \centering
    \begin{minipage}[b]{0.5\textwidth}
	    \centering
	    \includegraphics[width=7cm, height=5cm]{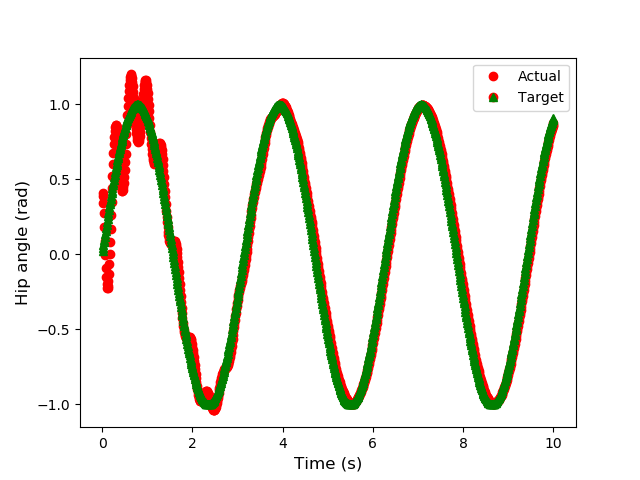}
	    \subcaption{Desired vs Actual Hip Angle over time.}
    \end{minipage}%
    \begin{minipage}[b]{0.5\textwidth}
        \centering
        \begin{tabular}{|c|c|}
            \hline
            \textbf{Parameter} & \textbf{Value} \\
            \hline
            $K_p$       &  22.5\\
            \hline
            $K_d$       &  0.85\\
            \hline
        \end{tabular}
        \subcaption{PD Gains}
    \end{minipage}%
    \centering
    \caption{"Pinned-to-the-ground” method hip controller performance and gains.}
    \label{fig:hip_pd_ground}
\end{figure}

In Section \ref{hip_pd_controller}, we have explained two different methods for tuning the PD gains for the hip controller. The different gains yielded by these methods and their performance to rotate at the desired sine-wave angle are shown in Figure \ref{fig:hip_pd_air} and \ref{fig:hip_pd_ground}.

In Figure \ref{fig:hip_pd_air}, the anomaly at the beginning of the graph is caused by the initial conditions of the hip at the start of the simulation, where it doesn't behave as it should. We can see though that, in both methods of tuning, the fluctuations of the hip angle stabilise after about $2.5s$ and the hip actuator performs as desired. 

\subsubsection{Knee PD Tuning Results}
\begin{figure}[ht]
    \centering
    \begin{minipage}[b]{0.5\textwidth}
	    \centering
	    \includegraphics[width=7cm, height=5cm]{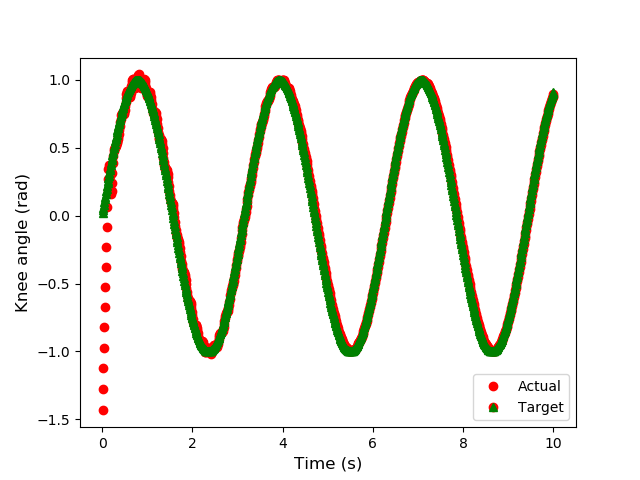}
	    \subcaption{Desired vs Actual Knee Angle over time.}
    \end{minipage}%
    \begin{minipage}[b]{0.5\textwidth}
        \centering
        \begin{tabular}{|c|c|}
            \hline
            \textbf{Parameter} & \textbf{Value} \\
            \hline
            $K_p$       &  200.0\\
            \hline
            $K_d$       &  4.0\\
            \hline
        \end{tabular}
        \subcaption{PD Gains}
    \end{minipage}%
    \centering
    \caption{Knee controller performance and gains.}
    \label{fig:knee_pd}
\end{figure}

In Section \ref{knee_pd_controller}, we have explained the upside-down method of tuning the gains for the knee PD controllers. The gains yielded and the method's performance to rotate at the desired sine-wave angle are shown in Figure \ref{fig:knee_pd}. The knee actuator rotates to the desired angle in a stable manner.
\subsubsection{Ankle PD Tuning Results}
\begin{figure}[ht]
    \centering
    \begin{minipage}[b]{0.5\textwidth}
	    \centering
	    \includegraphics[width=7cm, height=5cm]{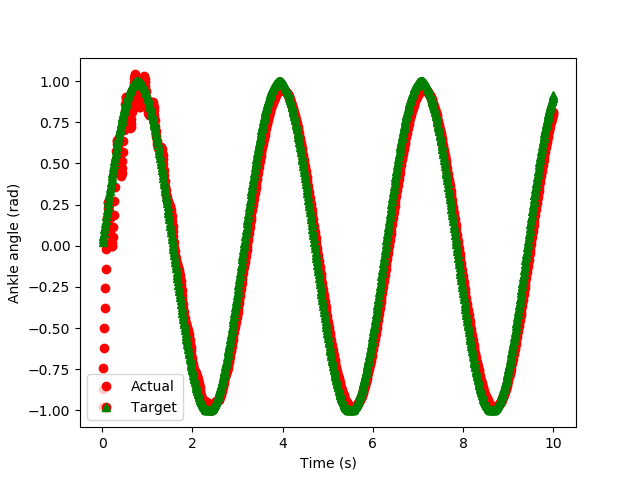}
	    \subcaption{Desired vs Actual Ankle Angle over time.}
    \end{minipage}%
    \begin{minipage}[b]{0.5\textwidth}
        \centering
        \begin{tabular}{|c|c|}
            \hline
            \textbf{Parameter} & \textbf{Value} \\
            \hline
            $K_p$       &  20.0\\
            \hline
            $K_d$       &  1.2\\
            \hline
        \end{tabular}
        \subcaption{PD Gains}
    \end{minipage}%
    \centering
    \caption{Ankle controller performance and gains.}
    \label{fig:ankle_pd}
\end{figure}

In Section \ref{ankle_pd_controller}, we have explained the upside-down method of tuning the gains for the ankle PD controllers. The gains yielded by this method and its performance to rotate at the desired sine-wave angle are shown in Figure \ref{fig:ankle_pd}. The performance of the Ankle controller is good, as after $2.5s$ the ankle actuator rotates to the desired angle in a stable manner.

\subsubsection{Upright Posture PD Controller}
This PD controller was implemented for the balance control of the robot, see Section \ref{balance_control}. It was tuned directly on the biped robot during walking. As discussed in Section \ref{upright_pd_section}, this PD controller is an adjustment controller, therefore the values for the PD gains should not be high, so that the hip PD controllers are not greatly affected. Therefore, during manual tuning and a lot of experimentation, we decided the ideal PD gains were $K_p = 1.5$ and $K_d = 0.1$. We also came to the conclusion that the desired torso angle $\theta_d$ should not be $0.0$, as we previously assumed was the ideal value. Having $\theta_d = 0.0$, wouldn't give the biped the necessary push to move forward, so in the approaches explained later on we are using different values as $\theta_d$, giving the torso of the biped a slight inclination to constantly "force" it to move forward. This constant force forward does not make the robot lose stability.
\\ \\
To conclude this section, the PD controllers are performing quite well and perform stably after about $2.5s$. 

\subsection{Biped with point feet}
As we mentioned in Section \ref{inital_simulation}, we use the robot structure of another student's Master's thesis \cite{Zeyu_project}, which uses a four-joint robot with point feet. In our attempt to replicate his work and expand, we found difficulty in achieving stable walking. He managed to do that, with the robot being able to do 12 steps and then falling due to jitter in the torso of the robot.
\begin{figure}[ht]
    \centering
    \includegraphics[width=\textwidth, height=6.6cm]{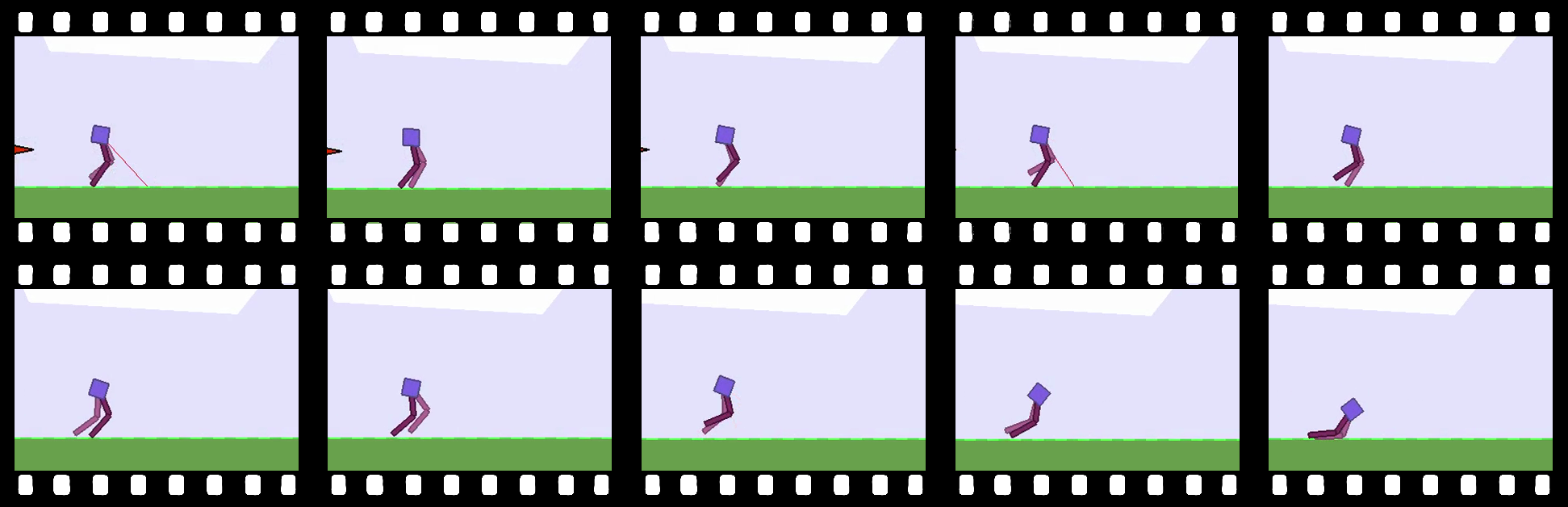}
    \caption{Snapshots of biped with point feet during simulation. Watch video here: \url{https://youtu.be/D-mHUl0IK-4}.}
    \label{fig:no_foot_snaps}
\end{figure}

Our implementation tried to overcome the jitter in the torso of the robot by adding balance control, see Section \ref{balance_control}, through an Upright Posture PD Controller, see Section \ref{upright_pd_section}. It's important to note that with desired torso angle $\theta_d = 0.0$, the robot would not do a single step, therefore we used $\theta_d = 0.3$ to give it a natural forward inclination in order to move in the desired forward direction. We managed to get the robot to walk four steps, but unable to form a cyclic pattern and achieve stability. 
\begin{figure}[ht]
    \centering
    \begin{minipage}[b]{0.5\textwidth}
	    \centering
	    \includegraphics[width=7cm, height=5cm]{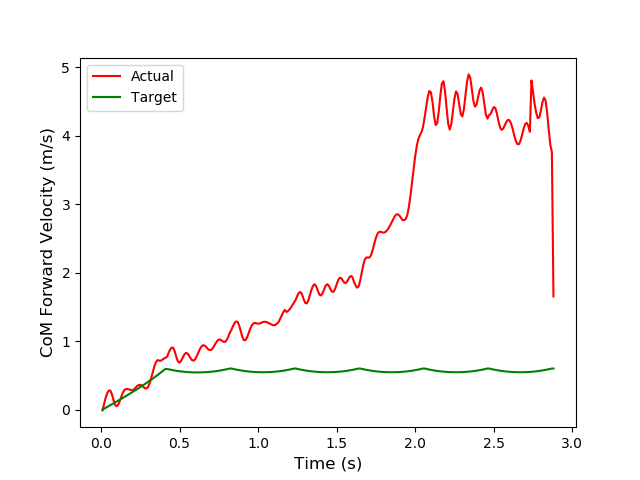}
	    \subcaption{Desired vs Actual Forward velocity of CoM.}
    \end{minipage}%
    \begin{minipage}[b]{0.5\textwidth}
        \centering
        \begin{tabular}{|c|c|c|}
            \hline
            \textbf{Parameter}  &  \textbf{Hip} & \textbf{Knee} \\
            \hline
            $K_p$               &  100.5        &   200.0\\
            \hline
            $K_d$               &  0.85         &   4.0\\
            \hline
        \end{tabular}
    \subcaption{PD gains for each joint.}
    \end{minipage}%
    \centering
    \caption{Results for biped robot with point feet.}
    \label{fig:no_feet}
\end{figure}

The parameter values for the PD controllers had to be fine-tuned in the environment for the walking task. Figure \ref{fig:no_feet} shows the forward velocity of the CoM against the desired one over time and the values we used for the control of the biped with point feet. Because of the instability, the velocity is increasing throughout the short locomotion of the biped up to the point it falls to the ground.

\subsection{Biped with feet}
After the failure of achieving stable walking with point feet, we decided to move on and add feet to the biped in order to make walking more stable but also more natural and human-like. Our six-joint biped went through three phases of evolution, outlined below.

\subsubsection{Feet with no ankle control}
In our first attempt with the robot with feet, we tried to achieve walking without implementing ankle control. 
\begin{figure}[ht]
    \centering
    \includegraphics[width=\textwidth, height=6.6cm]{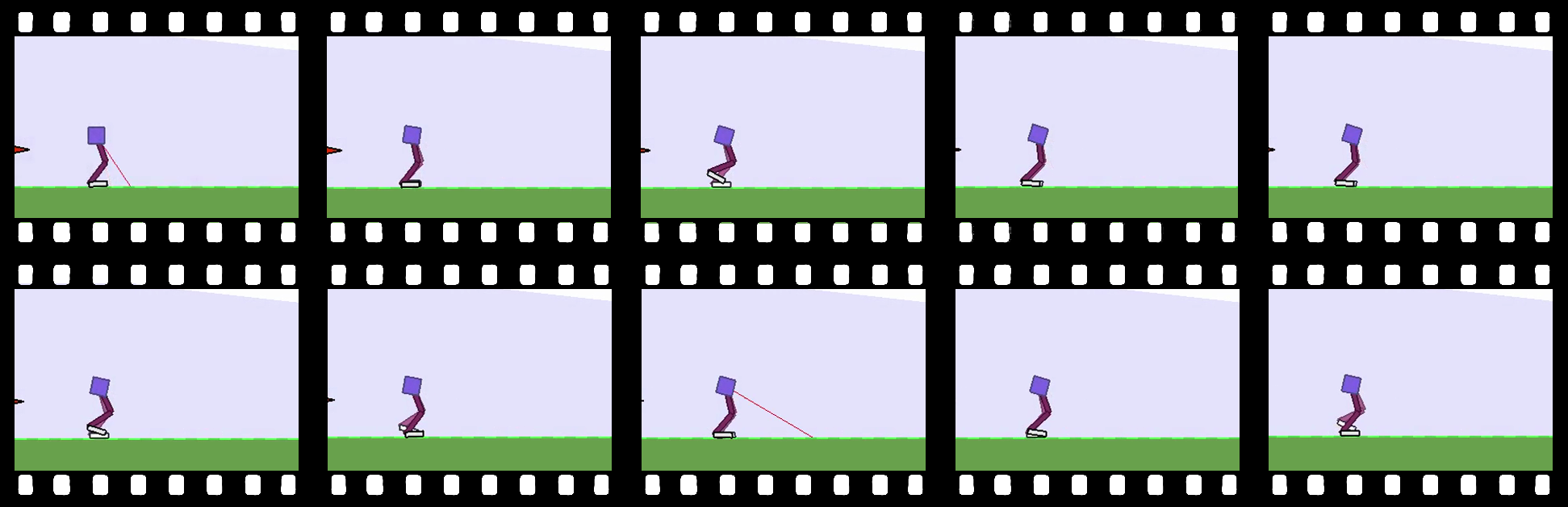}
    \caption{Snapshots of biped with feet but no ankle control during simulation. Watch video here: \url{https://youtu.be/tMFJ9AqREpM}.}
    \label{fig:no_ankle_snaps}
\end{figure}

The idea was to keep the feet at a constant angle parallel to the ground, which was decided to be $0.71 \, radians$ with respect to the shin, and with the LIPM and PTA for foot trajectory, the feet would be able to move high enough above the ground, such that the feet would simply act as a larger contact point during locomotion. 

\begin{figure}[ht]
    \centering
    \begin{minipage}[b]{0.5\textwidth}
	    \centering
	    \includegraphics[width=7cm, height=5cm]{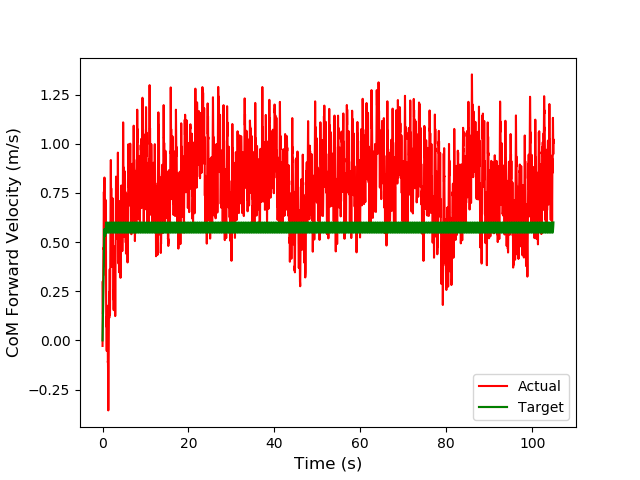}
	    \subcaption{Desired vs Actual Forward velocity of CoM.}
    \end{minipage}%
    \begin{minipage}[b]{0.5\textwidth}
        \centering
        \begin{tabular}{|c|c|c|c|}
            \hline
            \textbf{Parameter}  &  \textbf{Hip} & \textbf{Knee} & \textbf{Ankle}\\
            \hline
            $K_p$               &  45.5        &   200.0        &   0.0\\
            \hline
            $K_d$               &  0.85         &   4.0         &   0.0\\
            \hline
        \end{tabular}
    \subcaption{PD gains for each joint.}
    \end{minipage}%
    \centering
    \caption{Results for biped robot with no ankle control.}
    \label{fig:feet_no_control}
\end{figure}

The Upright Posture PD Controller implemented performed well, with $\theta_d = 0.2$ $radians$. The gains for this PD controller, as well as the knee PD controller, were kept the same. The hip PD controller had to be fine-tuned. Figure \ref{fig:feet_no_control} shows the forward velocity of the CoM against the desired one over time and the values we used for the control of the biped with feet but no ankle control.

\begin{figure}[ht]
    \includegraphics[width=8cm,height=6cm]{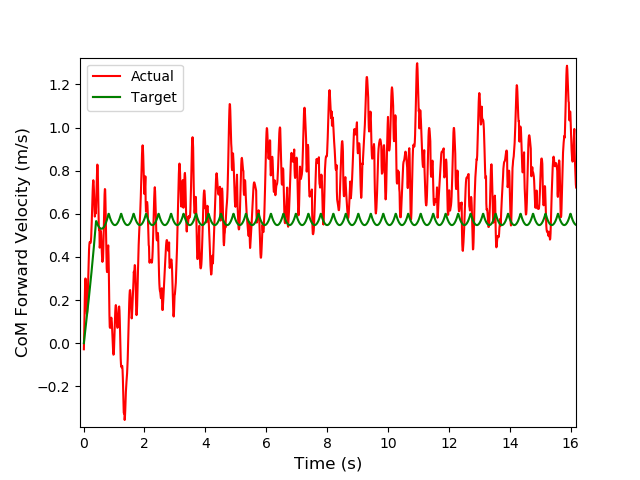}
    \centering
    \caption{Desired vs Actual Forward velocity of CoM at start of simulation.}
    \label{fig:feet_no_control_zoom}
\end{figure}
At the start of the simulation, as can be seen in first snapshots in Figure \ref{fig:no_ankle_snaps}, the biped is doing seven steps where it seems to lean too much in the forward direction and then because of the restriction of the LIPM of constant CoM height, it tries to correct the error. In that attempt, it over-corrects itself and then leans forward again to bring its CoM at the desired constant height. This fluctuation can be seen closer in Figure \ref{fig:feet_no_control_zoom}, between $0-4s$. After the seven steps, the biped walks the whole terrain of the simulation in a cyclic pattern stably, doing an overall of 256 steps. After those many steps, we assume the biped can keep walking for many more.

\subsubsection{Feet with "wrong" ankle control}
\begin{figure}[ht]
    \centering
    \begin{minipage}[b]{0.5\textwidth}
	    \centering
	    \includegraphics[width=7cm, height=5cm]{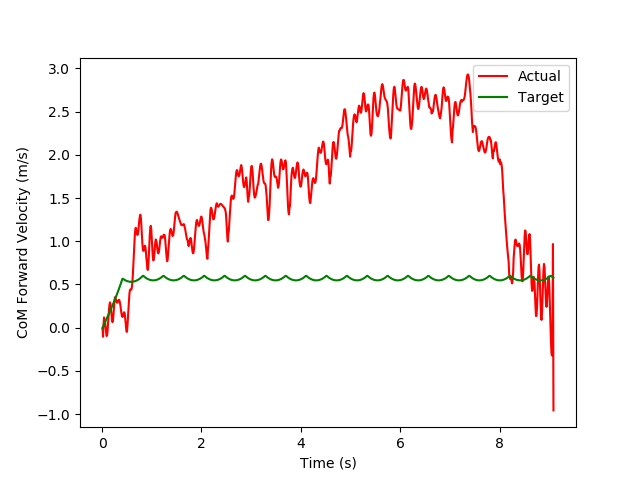}
	    \subcaption{Desired vs Actual Forward velocity of CoM.}
    \end{minipage}%
    \begin{minipage}[b]{0.5\textwidth}
        \centering
        \begin{tabular}{|c|c|c|c|}
            \hline
            \textbf{Parameter}  &  \textbf{Hip} & \textbf{Knee} & \textbf{Ankle}\\
            \hline
            $K_p$               &  45.5        &   200.0       &   20.0\\
            \hline
            $K_d$               &  0.85         &   4.0         &   1.2\\
            \hline
        \end{tabular}
    \subcaption{PD gains for each joint.}
    \end{minipage}%
    \centering
    \caption{Results for biped robot with "wrong" ankle control.}
    \label{fig:feet_wrong_control}
\end{figure}

The second attempt with the robot with feet was to implement proper ankle controllers to control the movement of the feet in accordance to the joint angles solutions, given by our Inverse Kinematics analysis, see Section \ref{section:IK}. A mistake in our calculations led to the robot to perform in a more running manner, as we can see from the increasing forward velocity in Figure \ref{fig:feet_wrong_control}.
\begin{figure}[ht]
    \centering
    \includegraphics[width=\textwidth, height=6.6cm]{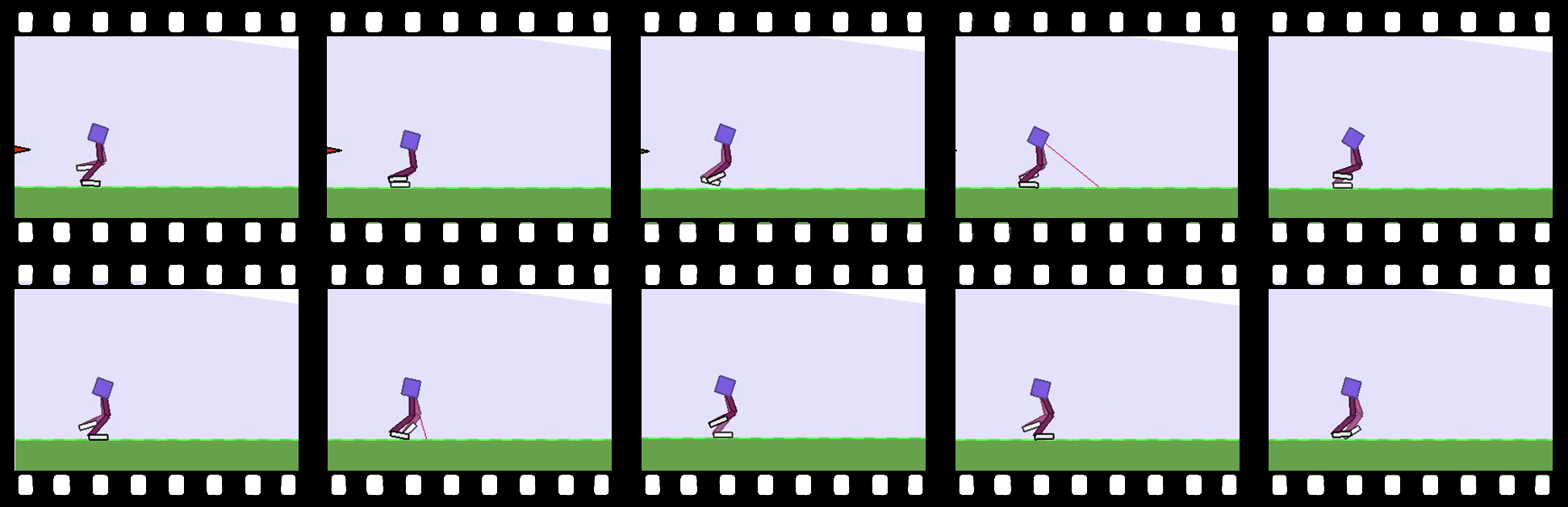}
    \caption{Snapshots of biped with feet and "wrong" ankle control during simulation. Watch video here: \url{https://youtu.be/X4IdAwhBIhY}.}
    \label{fig:wrong_ankle_snaps}
\end{figure}

Despite the mistake, we believe our results were worth mentioning as the robot ran 21 steps before falling to the ground. The feet acted in a "flappy" manner, see Figure \ref{fig:wrong_ankle_snaps}, making the biped to have a slight jump in every step, causing the increase in forward velocity. Also, our Upright Posture PD controller's desired torso angle was set at $\theta_d = 0.3$ and this inclination is another factor that boosted the increase. The PD gains of all controllers were kept the same as the previous version of the biped with no ankle control.

\subsubsection{Feet with correct ankle control}\label{section:correct_ankle_control}
In the final and most successful version of our biped, we corrected our Inverse Kinematics calculations and the robot performs in a cyclic pattern and shows stable walking that looks natural as well. 
\begin{figure}[ht]
    \centering
    \includegraphics[width=\textwidth, height=6.6cm]{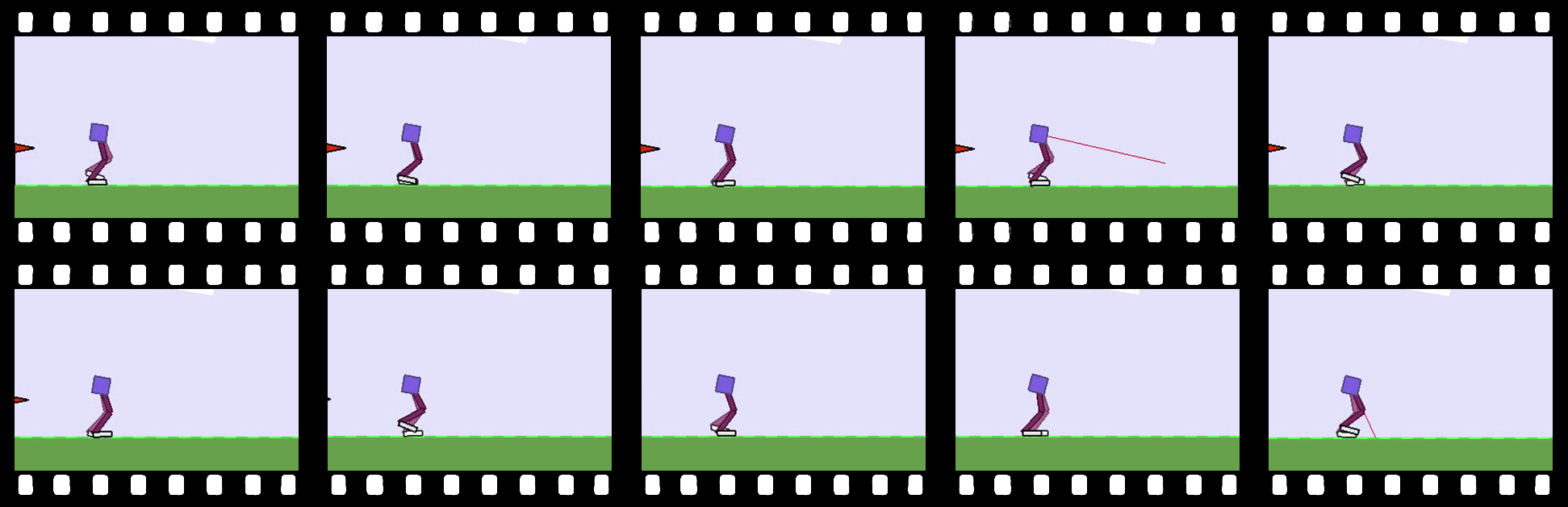}
    \caption{Snapshots of biped with feet and correct ankle control during simulation. Watch video here: \url{https://youtu.be/yABq_4JU8bg}.}
    \label{fig:correct_ankle_snaps}
\end{figure}
The biped walks at a velocity fluctuating around $1.0m/s$, reaching maximums of $1.5m/s$ in some periods. 

\begin{figure}[ht]
    \centering
    \begin{subfigure}[t]{0.5\textwidth}
	    \centering
	    \includegraphics[width=7cm, height=5cm]{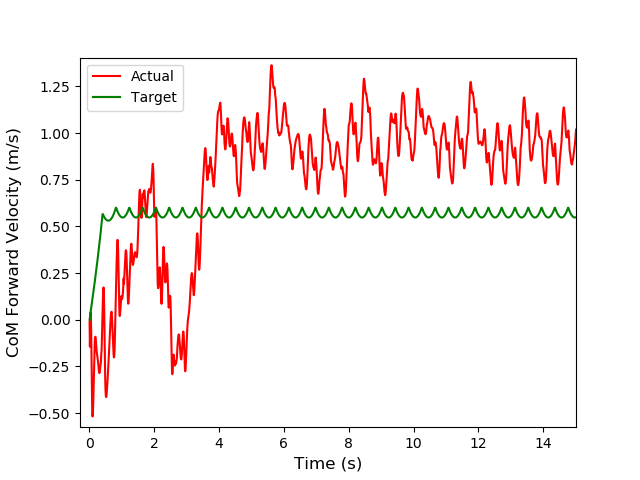}
	    \caption{}
	    \label{first_xi}
    \end{subfigure}%
    \begin{subfigure}[t]{0.5\textwidth}
	    \centering
	    \includegraphics[width=7cm, height=5cm]{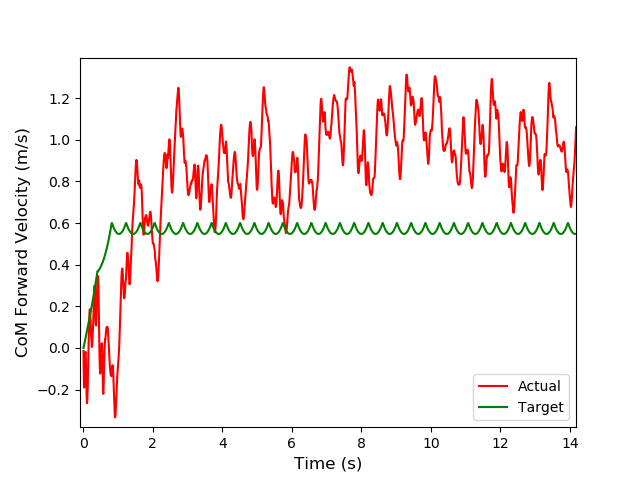}
	    \caption{}
	    \label{second_xi}
    \end{subfigure}
    \centering
    \caption{Comparison of Forward velocity of CoM, changing its initial position $x_i$ on x-axis: (a) $x_i = +0.25$, (b) $x_i = +0.173$.}
    \label{fig:compare_initial_positions}
\end{figure}

The same problem at the start of the simulation occurred in this phase as well, where the robot would lean too much forward, then lean back over-correcting itself and then getting to a balanced height to proceed to stable walking. In order to tackle this, we tried to vary the initial position of the CoM in the x-axis, which affected the LIPM calculations and the PTA for foot placement. Figure \ref{fig:compare_initial_positions} shows the forward velocity of the CoM with different initial positions of the CoM.
Graph (a) shows the initial position $x_i = +0.25$, where the biped does eleven steps (leaning forwards and backwards) in order to get into a stable position and walks an overall 210 steps to 
complete the whole terrain ground of the simulation. Graph (b) shows $x_i = +0.173$, where the biped only does five steps, \textit{not} leaning forwards and backwards but doing steps on the
\begin{figure}[ht]
    \centering
    \begin{minipage}[b]{0.5\textwidth}
	    \centering
	    \includegraphics[width=7cm, height=5cm]{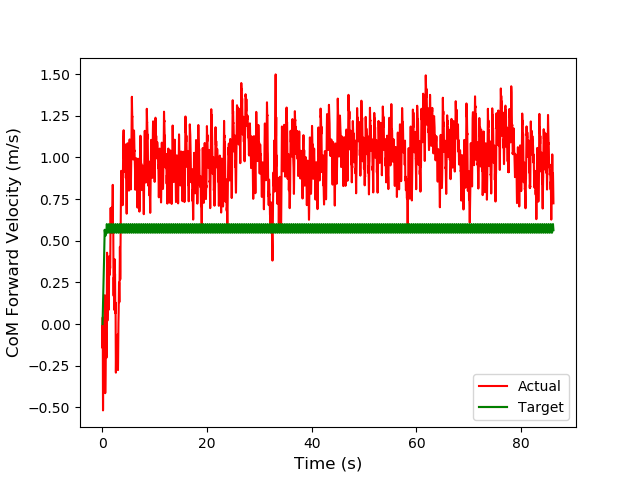}
	    \subcaption{Desired vs Actual Forward velocity of CoM.}
    \end{minipage}%
    \begin{minipage}[b]{0.5\textwidth}
        \centering
        \begin{tabular}{|c|c|c|c|}
            \hline
            \textbf{Parameter}  &  \textbf{Hip} & \textbf{Knee} & \textbf{Ankle}\\
            \hline
            $K_p$               &  48.5        &   200.0       &   20.0\\
            \hline
            $K_d$               &  0.85         &   4.0         &   1.2\\
            \hline
        \end{tabular}
    \subcaption{PD gains for each joint.}
    \end{minipage}%
    \centering
    \caption{Results for biped robot with correct ankle control.}
    \label{fig:feet_correct_control}
\end{figure}
spot, stabilising and going forward until the end of the terrain ground, completing it in 203 steps. This makes sense, as the biped takes less steps to get into a stable situation. Figure \ref{fig:feet_correct_control} shows the final results of our implementation of biped walking, choosing $x_i = +0.173$.

Our Upright Posture PD Controller performs well to keep $\theta$ near $\theta_d = 0.1 \: radians$, which is the closest to the ideal value of $0.0 \: radians$ we could get. The hip PD controller's $K_p$ value was increased slightly to $48.5$ in this version. This goes to show how significant fine-tuning of the PD gains is to the performance of our controllers. 
\begin{figure}[ht]
    \centering
    \begin{minipage}[b]{0.5\textwidth}
	    \centering
	    \includegraphics[width=7cm, height=5cm]{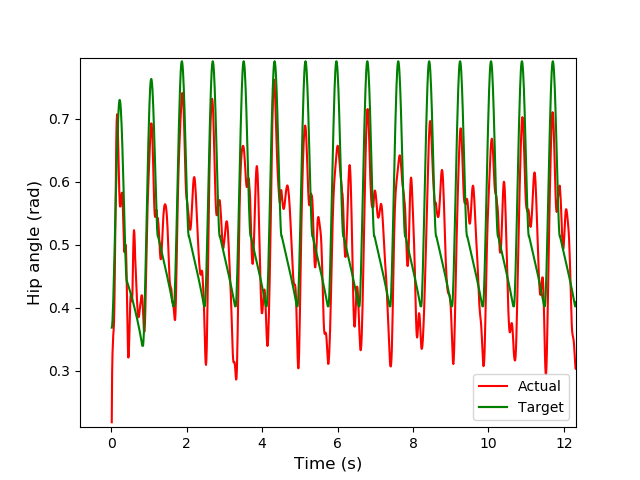}
	    \subcaption{Hip angle comparison.}
    \end{minipage}%
    \begin{minipage}[b]{0.5\textwidth}
	    \centering
	    \includegraphics[width=7cm, height=5cm]{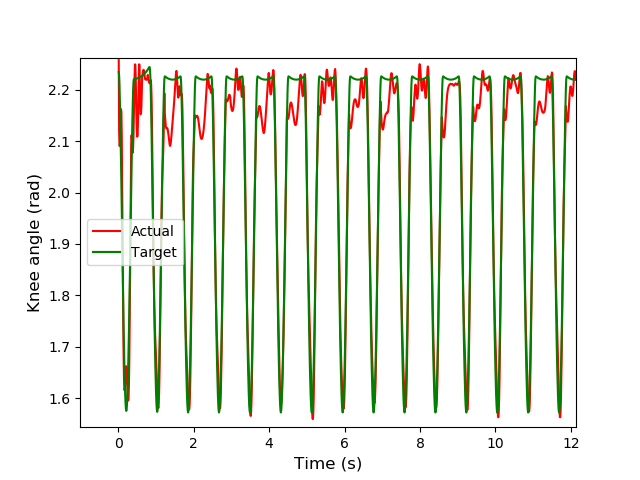}
	    \subcaption{Knee angle comparison.}
    \end{minipage}%
    \centering
    \caption{Comparison of Desired angle vs Actual angle produced by PD Controllers.}
    \label{fig:joints_performance}
\end{figure}

It is worthwhile to show the joint angles produced by the biped's actuators, at least for our last and successful implementation. Figure \ref{fig:joints_performance} shows the desired angles calculated by our Inverse Kinematics analysis, against the actual angles produced by the joints actuators, decided by our PD controllers.
\newpage

\chapter{Conclusion}
In this project, a bipedal walking control algorithm based on LIPM and the three-part intuitive controller by Raibert is proposed. Controlling the height, forward velocity and balance of a bipedal robot yields a stable walking gait. The relationship between foot placement and forward velocity becomes linear with proper control of height and balance. Our simplified robot structure allowed the use of the LIPM for the motion of the CoM, a simple model, which we combined with the PTA algorithm for the trajectory of the swing leg. 

The use of Inverse Kinematics analysis allowed us to control CoM position or foot placement of the biped by controlling the joint angles. In order to ensure the torques applied by the joint motors will provide the desired angles, PD controllers are implemented for each joint. A separate PD controller is implemented for balance control. A low-pass filter is applied on the measured velocity to reduce high-frequency signal mutations. 

Our experimentation in the simulated environment went through a couple of phases of "evolution", before reaching the final robot structure and a stable dynamic walking gait which resulted in an almost natural human-like walk. It was a journey from a biped with point feet, performing unstably, to a biped with feet with proper six-joint control.

Comparing the last version with the robot with no ankle control, we can see that it walks faster and thus does less steps to complete the same distance. The walking has a cyclic pattern and is stable in both versions, but the latest one is more natural and also more mathematically correct. 

As we mentioned in our results, the forward velocity of the CoM is always above the desired velocity. One reason for this is the constant inclination of the torso in the forward direction. This causes the whole body to lean slightly forward. The joint angles calculated from the Inverse Kinematics analysis, will be slightly "tilted" towards that inclination and that enhances the increased velocity. The constant height constraint of the LIPM ensures that, despite the inclination forward, the robot "fixes" itself backwards. This, along with our assumption that ground contact does not affect the motion of our system, causes the fluctuations in the forward velocity as shown in the previous chapter.

Our results show that simple systems, like the LIPM and intuitive controllers, can achieve stable dynamic walking gait on flat ground, with the potential of non-stop walking. However, we do not consider unexpected obstacles or external forces in our project, thus our control is not robust in that sense. Another issue we were unable to resolve is the fact that the forward velocity of the CoM is always above the desired velocity. It is important to be able to control the overall performance of any system.

Future work can be carried out to solve these issues. Research can be done to tackle unexpected external forces that could force the robot off balance. The edge of the foot is an underactuated DOF and will act as a pivot if the foot tilts. Various methods have been proposed for push recovery, like Li et al.'s ankle torque control and body attitude control \cite{DBLP:journals/trob/LiZZX17}. Another problem that affects the LIPM trajectories is foot slipping, which can be reduced with additional feedback control. Fujimoto et al. proposed a feedback compensation of yaw axis rotation by arm swing motion \cite{arm_motion}. Finally, due to time constraints enforced on the project, we were unable to apply the online linear regression analysis for foot placement to the LIPM \cite{DBLP:conf/iros/YouLCT15}. This can improve the robustness of our control system, as coefficients in the foot placement estimators are replaced with online parameters based on previous steps data.

\newpage

\bibliographystyle{unsrt}

\end{document}